\documentclass{ieeeaccess}
\usepackage{dirtree}
\usepackage{cite}
\usepackage{amsmath,amssymb,amsfonts} % matched
\usepackage{algorithmic}
\usepackage{bm}                   % matched
\usepackage{times}                % matched
\usepackage{epsfig}               % matched
\usepackage{float}                % matched
\floatstyle{plaintop}
\restylefloat{table}
\usepackage{array}                % matched
\usepackage{multicol}             % matched
\usepackage{multirow}             % matched
\usepackage{graphicx}            % matched
\newcolumntype{x}[1]{>{\centering\arraybackslash\hspace{0pt}}p{#1}}
\usepackage{url}
\usepackage{pifont}              % matched
\usepackage[symbol]{footmisc}    % matched
\usepackage{booktabs}

\usepackage{textcomp}
\def\BibTeX{{\rm B\kern-.05em{\sc i\kern-.025em b}\kern-.08em
    T\kern-.1667em\lower.7ex\hbox{E}\kern-.125emX}}
\begin{document}
\history{Date of publication September 7, 2021, date of current version September 17, 2021.}
\doi{10.1109/ACCESS.2021.3110973}

\title{ABN: Agent-Aware Boundary Networks \\ for Temporal Action Proposal Generation}
\author{\uppercase{Khoa Vo}\authorrefmark{1},
\uppercase{Kashu~Yamazaki}\authorrefmark{1}, 
\uppercase{Sang~Truong}\authorrefmark{1}, 
\uppercase{Minh-Triet~Tran}\authorrefmark{3},
\uppercase{Akihiro~Sugimoto}\authorrefmark{2}, 
\IEEEmembership{Member, IEEE}, 
\uppercase{Ngan Le}\authorrefmark{1}, \IEEEmembership{Member, IEEE}}
\address[1]{University of Arkansas, Fayetteville, AR 72703 (e-mail: \{khoavoho, thile, kyamazak, sangt\}@uark.edu)}
\address[2]{National Institute of Informatics, Japan (e-mail: sugimoto@nii.ac.jp)}
\address[3]{University of Science - VNU-HCM, Vietnam (e-mail: tmtriet@hcmus.edu.vn)}
\tfootnote{
This work was supported by the National Science Foundation under Award OIA-1946391; funded by Gia Lam Urban Development and Investment Company Limited, Vingroup, and partially supported by Vingroup Innovation Foundation (VINIF) under project code VINIF.2019.DA19. \\
Any opinions, findings, and conclusions or recommendations expressed in this material are those of the author(s) and do not necessarily reflect the views of the National Science Foundation.
}

\markboth
{Khoa Vo \headeretal: Preparation of Papers for Journal of IEEE Access}
{Khoa Vo \headeretal: Preparation of Papers for Journal of IEEE Access}

\corresp{Corresponding author: Khoa Vo (e-mail: khoavoho@uark.edu).}

\begin{abstract}
Temporal action proposal generation (TAPG) aims to estimate temporal intervals of actions in untrimmed videos, which is a challenging yet plays an important role in many tasks of video analysis and understanding. Despite the great achievement in TAPG, most existing works ignore the human perception of interaction between agents and the surrounding environment by applying a deep learning model as a black-box to the untrimmed videos to extract video visual representation. Therefore, it is beneficial and potentially improve the performance of TAPG if we can capture these interactions between agents and the environment. In this paper, we propose a novel framework named Agent-Aware Boundary Network (ABN), which consists of two sub-networks (i) an Agent-Aware Representation Network to obtain both agent-agent and agents-environment relationships in the video representation, and (ii) a Boundary Generation Network to estimate the confidence score of temporal intervals. In the Agent-Aware Representation Network, the interactions between agents are expressed through local pathway, which operates at a local level to focus on the motions of agents whereas the overall perception of the surroundings are expressed through global pathway, which operates at a global level to perceive the effects of agents-environment. Comprehensive evaluations on 20-action THUMOS-14 and 200-action ActivityNet-1.3 datasets with different backbone networks (i.e C3D, SlowFast and Two-Stream) show that our proposed ABN robustly outperforms state-of-the-art methods regardless of the employed backbone network on TAPG. We further examine the proposal quality by leveraging proposals generated by our method onto temporal action detection (TAD) frameworks and evaluate their detection performances. The source code can be found in this URL \footnote{https://github.com/vhvkhoa/TAPG-AgentEnvNetwork.git}.
\end{abstract}

\begin{keywords}
Temporal Action Proposal Generation, Temporal Action Detection, Agent-Aware Boundary Network
\end{keywords}

\titlepgskip=-15pt

\maketitle

\section{Introduction}
\label{sec:introduction}

% + ADD AN EXAMPLE WHERE WE CAN SEE THAT AGENT AND ENVIRONMENT EFFECT TO THE ACTION PROPOSAL GENERATION
%  + NONE OF THE PREVIOUS WORK HAS EXPLOITED AND UTILIZED THE RELATIONSHIP BETWEEN AGENT WHO PERFORMS ACTION AND ENVIRONMENT WHICH RECEIVE THE ACTION
 
%{Submit to IEEE Access}

Temporal action proposal generation (TAPG) \cite{SST_CVPR2017, CTAP, TCN, MSRA, actionproposal_2016, bmn, anchor_1, anchor_2, anchor_3, FasterR_CNN_Action, boundary_0, lin2018bsn, liu2019multi, dbg} is one of the most key and fundamental tasks in video understanding i.e.  action recognition \cite{2_stream_1, SlowFast}, video summarization \cite{summarization_2015, summarization_2016}, video captioning \cite{video_captioning_cvpr2020, captioning_2020}, video recommendation \cite{video_recommendation_1}, video highlight detection \cite{highlight_detection_CVPR19}, and smart surveillance \cite{smart_surveillance_1, smart_surveillance_2}. Given an untrimmed video, TAPG aims to propose temporal intervals with specific starting and ending timestamps for each action. Most of existing TAPG approaches first detect a set of possible starting and ending timestamps of all actions separately, and then a proposal evaluation module is employed to evaluate every possible pair of starting and ending timestamps by predicting its confidence score. The non-maximum suppression (NMS) function is finally used to eliminate redundant candidate proposals based on their confidence scores and overlapping metrics. A robust TAPG method should be able to (i) generate temporal proposals with actual boundaries to cover action instances precisely and exhaustively; (ii) cover multi-duration actions; (iii) generate reliable confidence scores so that proposals can be retrieved properly \cite{bmn}. Despite recent endeavors \cite{lin2018bsn, bmn, liu2019multi}, TAPG remains as an open problem, especially when facing real-world problems such as action duration variability, activity complexity, camera motion, viewpoint changes, etc.
In spite of good achievements on benchmarking datasets, the existing TAPG approaches have some limitations as follows:

\begin{itemize}
\item In previous works, video visual representation is extracted by directly applying a backbone model, e.g. C3D/I3D \mbox{\cite{C3D_3}}, Two-Stream network \mbox{\cite{2_stream_1, 2_stream_2}} or SlowFast network \mbox{\cite{SlowFast}} into the whole spatial dimensions of the video (or entire snippet). This makes the predictions biased to the background (or environment) instead of agents who commit actions because the agents with their actions usually occupy a small region compared to the entire frame.
\item A temporal action proposal is a combination of three entities i.e.  agent, action, and environment; however, the existing approaches do not have any mechanism to present such combination as well as express the relationship between these entities.
%\item Most of the existing approaches are unable to follow the human perceiving process of understanding the video content. As the human perceiving process of understanding, a person focuses on deciding what an agent is doing by observing the information from both agent activities and the environment around the agent. Such a human perceiving process is not presented in any previous works which apply a backbone network into entire spatial dimensions of video snippets of frames (e.g. 8-frame snippets or 16-frame snippets, etc.)}
\end{itemize}

\begin{figure*}[ht]
  \begin{center}
    \includegraphics[width=0.95\linewidth]{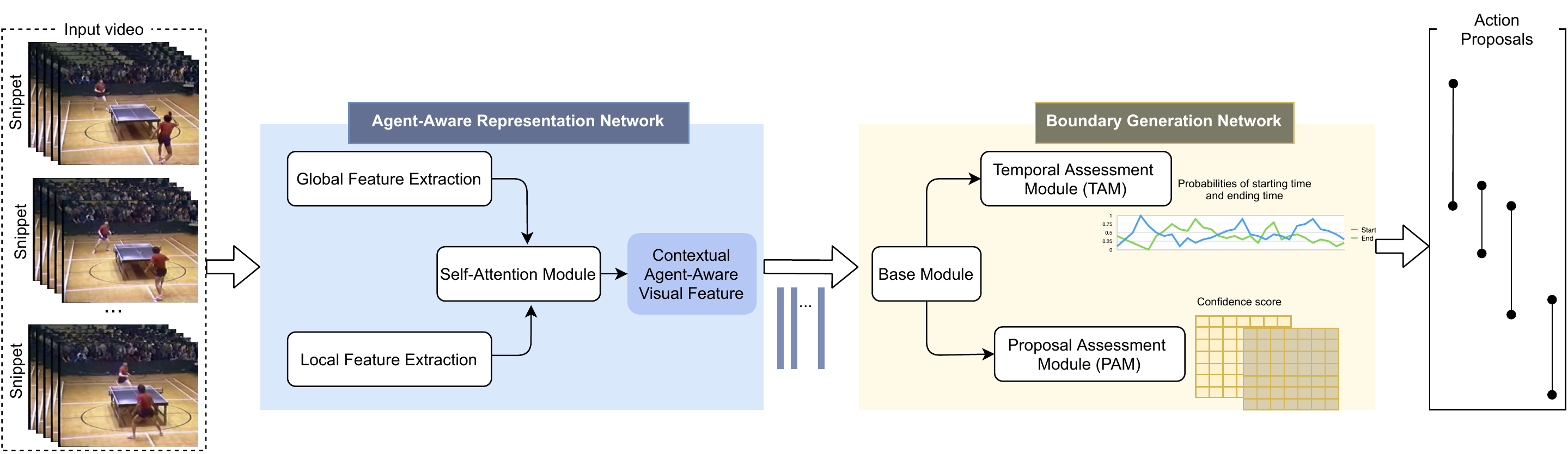}
    \caption{An overview architecture for our proposed Contextual Agent-Aware Boundary Network (ABN) for TAPG. ABN consists of two main sub-networks i.e. \textit{Agent-Aware Representation Network} and \textit{Boundary Generation Network}.}
    \label{fig:overall}
  \end{center}
  %\vspace*{-\baselineskip}
\end{figure*}

To address the aforementioned limitations, we leverage human perception process of a temporal action proposal which is a combination of three entities i.e.  agent, action, and environment and we propose a novel \textbf{Contextual Agent-Aware Boundary Network (ABN)}. Our ABN consists of two main sub-networks i.e. \textit{Agent-Aware Representation Network} and \textit{Boundary Generation Network}. The first sub-network aims to extract video visual representation i.e. contextual agent-aware visual feature, given an untrimmed video whereas the second sub-network aims to estimate the confidence score matrices and the probabilities of starting time and ending time given a video feature. To interpret those entities of agent, action, and environment, our Agent-Aware Representation Network comprises of two semantic pathways corresponding to \textit{local pathway}, which locally extracts information from the agents who commit actions and \textit{global pathway}, which globally extracts information from entire environment. Furthermore, the number of agents in a given video can be arbitrary; however, a few of them are actually committing the action. To extract a semantic local feature, we apply a self-attention module. The final video feature combines both local feature and global feature through a self-attention module. The second sub-network, Boundary Generation Network, takes contextual agent-aware visual feature as an input and consists of three modules corresponding to Base Module to model the temporal relationship as well as provide a shared feature sequence for later modules of Temporal Assessment Module (TAM) and Proposal Assessment Module(PAM). The overall flowchart of our proposed ABN is given in Fig.\mbox{\ref{fig:overall}}.

\textbf{Our main contributions are summarized as follows:}
\begin{enumerate}
    \item  Leveraging the human perception process of understanding an action which combines agents, action and environment, we propose an end-to-end contextual ABN for TAPG. Our ABN contains two sub-networks corresponding to (i) Agent-Aware Representation Network for extracting semantic video feature given untrimmed video and (ii) Boundary Generation Network to evaluate confidence scores of densely distributed proposals.
    \item Introducing Agent-Aware Representation Network, a novel video contextual visual representation, for extracting video feature from an untrimmed video. Our semantic  Agent-Aware Representation Network involves two parallel pathways: (i) local pathway to tell what agents are doing (ii) global pathway to express the relationship between the agents and the environment.
    \item Investigating the impacts of agents and the environment as well as the interaction between agents and their environment in our proposed ABN framework.
    \item Examining the action proposal quality and effectiveness of our proposed ABN by putting proposals that it generated to TAD framework and evaluate its detection performance.
    \item Benchmarking the proposed ABN on popular datasets in both TAPG and TAD, namely ActivityNet-1.3 with three different backbone networks (i.e. C3D, SlowFast and Two-Stream) and THUMOS-14 with two backbone networks (i.e. C3D and Two-Stream). Our proposed ABN has achieved state-of-the-art performance on both TAPG and TAD regardless of backbone network.
\end{enumerate}

\section{Related Work}

\subsection{Temporal Action Proposal Generation (TAPG)}
TAPG aims to propose temporal intervals that may contain an action instance with their temporal boundaries and confidence in untrimmed videos. In general, TAPG can be categorized into three groups i.e. anchor-based and boundary-based and hybrid anchor-boundary-based as follows:

\begin{itemize}
\item The anchor-based TAPG methods \cite{actionproposal_2016, FasterR_CNN_Action, anchor_1, anchor_2, anchor_3, SST_CVPR2017, anchor_5, CTAP} refer to the temporal boundary refinements of pre-defined anchors or sliding windows. Those methods are inspired by the achievements of anchor-based object detectors in still images like Faster R-CNN \cite{FasterRCNN}, RetinaNet \cite{RetinaNet}, or YOLOv3 \cite{yolov3}. These methods discretize the proposal task into a classification task where multiple predefined anchors with different lengths are used as classes and a class that best fits the ground truth action length is regarded as ground truth true class for training. In such approaches, a large number of proposals are densely generated. Although this approach helps to save computational costs, this approach lacks the flexibility of action duration.

\item The boundary-based TAPG methods \cite{boundary_0, lin2018bsn, liu2019multi} resolve the above problem by breaking every action interval into starting and ending points and learn to predict them. In those methods, there are two stages corresponding to generating the boundary probability sequence and applying the Boundary Matching mechanism to generate candidate proposals. In inference time, starting and ending probabilities at every time in the given video are predicted, then, those with local peaks will be chosen as potential boundaries. The potential starting points are paired with potential ending points to become a potential action interval when their interval fits in the predefined upper and lower threshold, along with a confidence score being a multiplication of the starting and ending probabilities. As one of the first boundary-based methods, \cite{boundary_0} defined actionness scores by grouping continuous high-score regions as the proposal. Later, boundary-sensitive method \cite{lin2018bsn} proposed a two-stage strategy where boundaries and actionness scores at every temporal point are predicted in the first stage and fused together, filtered by Soft-NMS to get the final proposals at the second stage. 

\item In order to make use of the advantages of both anchor-based methods and boundary-based methods, \cite{bmn, dbg} proposed hybrid approaches in which the boundary detection and the dense confidence regression are performed simultaneously by using ROI align. Based on the observation that anchor-based methods can uniformly cover all segments in videos but imprecise while boundary-based methods may have more precise boundaries but it may omit some proposals when the quality of actionness score is low, \cite{CTAP} proposed Complementary Temporal Action Proposal (CTAP) generator. BMN \cite{bmn} is an improvement of BSN \cite{lin2018bsn}. In BMN, a boundary-matching matrix is generated instead of actionness scores to capture an action-duration score for more descriptive final scores, which help to improve the final proposals' prediction. Continuously, drawing the inspiration from BSN \cite{lin2018bsn}, \cite{dbg} proposed Dense Boundary Generator (DBG) and implemented the boundary classification and action completeness regression for densely distributed proposals.
\end{itemize}

\begin{figure}[ht]
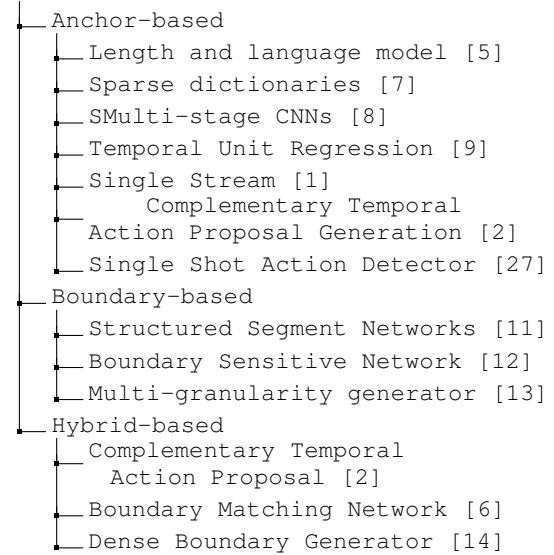

% \centering
\small
\dirtree{%
.1 Temporal Action Proposal Generation (TAPG).
.2 Anchor-based.
.3 Length and language model \cite{actionproposal_2016}.
.3 Sparse dictionaries \cite{anchor_1}.
.3 SMulti-stage CNNs \cite{anchor_2}.
.3 Temporal Unit Regression \cite{anchor_3}.
.3 Single Stream \cite{SST_CVPR2017}.
.3 \shortstack{Complementary Temporal \\Action Proposal Generation \cite{CTAP}}.
.3 Single Shot Action Detector \cite{anchor_5}.
.2 Boundary-based.
.3 Structured Segment Networks \cite{boundary_0}.
.3 Boundary Sensitive Network \cite{lin2018bsn}.
.3 Multi-granularity generator \cite{liu2019multi}.
.2 Hybrid-based.
.3 \shortstack{Complementary Temporal \\ Action Proposal \cite{CTAP}}.
.3 Boundary Matching Network \cite{bmn}.
.3 Dense Boundary Generator \cite{dbg}.
}

\caption{Approaches summarization on TAPG.}
\label{fig:fig_summary}
\end{figure}

The TAPG approaches can be summarized in Fig.~\ref{fig:fig_summary}.

% \subsection{Human in Video Understanding}

% Action Transformer~\cite{girdhar2019video} is a anchor-based approach which contains two subnetworks, i.e. base and head networks. It learns to attend to relevant regions of the person of interest and their context (other people, objects) to recognize the actions they are doing. Each head computes a clip embedding, which is used to focus on different parts like the face, hands and the other people to recognize that the person of interest is ‘holding hands’ and ‘watching a person’.

\subsection{Temporal Action Detection (TAD)}
Depending on spatial or temporal domain, action detection approaches can be categorised into either TAD (TAD) or spatial action detection (SAD) or spatial-temporal action detection. TAD aims to find the temporal intervals of starting action and ending action whereas SAD searches for human region and the corresponding human action in spatial domain. In this work, we focus on on TAD which provides the answer of what and when the action happens in a video.

Due to action recognition is a part of TAD, thus, most of the early TAD methods were built based on hand-crafted features, the same as action recognition. Early TAD methods are based on efficient spatio-temporal feature representations and motion propagation across frames in videos such as HOG3D~\mbox{\cite{klaser2008spatio}}, SIFT3D~\mbox{\cite{scovanner20073}}, ESURF~\mbox{\cite{willems2008efficient}}, MBH~\mbox{\cite{dalal2006human}} etc. As the performance of the methods using hand-crafted features became stabilized, TAD has reached a levelling off. With the convolutional neural networks (CNNs) was developed \mbox{\cite{krizhevsky2012imagenet}}, a lot of effective TAD approaches have proposed. In general, TAD can be divided into either one-stage detection or two-stage detection.

In one-stage framework, both temporal proposal and action classification are learnt simultaneously. Due to the similarity between TAD and object detection example, SSAD \mbox{\cite{anchor_5}}, SS-TAD \mbox{\cite{SSTAD_BMVC17}} made use of single-shot detector to solve TAD with one-stage detection. While both SSAD and SS-TAD make use of C3D feature \mbox{\cite{C3D_2, C3D_3, c3D_4, c3D_5}}, SS-TAD adopts the anchor mechanism and the stacked GRU units.

Unlike one-stage framework, two-stage approach is based on the paradigm of proposal generation-and-then classification i.e. extracts temporal proposals first, and then processes with the classification and regression operations. Similar to proposal generation in object detection, TAPG plays the most important role in TAD in this two-stage approach paradigm. Two-stage framework is the mainstream method, so most papers adopt this. TAD can be implemented by: (i) sliding windows such as S-CNN \mbox{\cite{anchor_2}} which fixes some size sliding windows to generate various sizes video segments, and then deal with them by a multi-stage network. S-CNN is build on C3D feature and contains thee sub-networks corresponding to TAPG, classification and localization; However, S-CNN is time consumption when increasing the overlap between the windows to obtains good performance, TURN \mbox{\cite{anchor_3}} leverages Faster RCNN to improe S-CNN by intergrating boundary regression network. (ii) boundary network such as BSN \mbox{\cite{lin2018bsn}}, BMN \mbox{\cite{bmn}}, DBG \mbox{\cite{dbg}} which aim to deal with video actions of different lengths and with precise temporal boundaries as well as reliable confidence scores. BSN first locates the boundaries of the temporal action segments i.e. starting time and ending time. Both starting time and ending time are then combined into temporal proposal. Based on the sequence of action confidence scores for each candidate proposal, a 32-dimensional proposal-level feature is extracted and benchmarked for evaluating the confidence of the temporal proposals. BMN and DBG are both improvements of BSN with a new confidence evaluation and boundary-matching mechanisms.

\subsection{Video Feature Representation}

Following the success of CNNs on image tasks. In~\cite{tran2015learning}, Tran et al. proposed a simple linear model named C3D which outperforms all previous best-reported methods. By transferring the 2D pre-trained model to 3D model, \cite{i3d_2017} proposed I3D. In I3D, the 3D filters are replaced by a set of repeated 2D filters. Inspired by the success of ResNet in image classification, Hara, et al. extended ResNet architecture to 3D CNN and proposed 3D ResNet \cite{resnet3D_50}. In their work, they examined various 3D CNN architecture under different backbone such as ResNet-18, ResNet-34, ResNet-50, ResNet-101, ResNet-152, ResNet-200, DenseNet-121 and ResNeXt-101.  The mainstream networks fall into three categories: Two-Stream networks, Recurrent Neural Network (RNN) with its popular variant named Long Short Term Memory (LSTM), and 3D networks. Two-Stream networks were first introduced by \cite{2_stream_1} and then they have been improved in \cite{2_stream_2}. Two-Stream networks explore video appearance and motion clues with two separate networks. One network is to exploit spatial information from individual frames while the other uses temporal information from optical flow. The outputs of the two networks are then combined by late fusion. RNN/LSTM is believed to cope with sequential information better, and thus many proposed methods \cite{LSTM_1, LSTM_2} attempted to incorporate LSTM to deal with action recognition. 3D networks, which were first introduced by \cite{C3D}, extract features from both the spatial and the temporal dimensions by performing 3D convolutions, thereby capturing the motion information encoded in multiple adjacent frames. Later on, C3D features, 3D CNN architectures and their improvements \cite{C3D_2, C3D_3, c3D_4, c3D_5, i3d_2017, resnet3D_50} have been proposed. In~\cite{tran2015learning}, Tran et al. proposed a simple linear model named C3D which outperforms all previous best-reported methods. By transferring the 2D pre-trained model to 3D model, \cite{i3d_2017} proposed I3D. In I3D, the 3D filters are replaced by a set of repeated 2D filters. Inspired by the success of ResNet in image classification, Hara, et al. extended ResNet architecture to 3D CNN and proposed 3D ResNet \cite{resnet3D_50}. In their work, they examined various 3D CNN architecture under different backbone such as ResNet-18, ResNet-34, ResNet-50, ResNet-101, ResNet-152, ResNet-200, DenseNet-121 and ResNeXt-101. Recently, SlowFast network \cite{SlowFast} is a variation of 3D CNN networks category, in which two parallel pathways are utilized to capture appearances of video scene and object motion in each pathway. SlowFast networks have been proposed to tackle the action recognition and action spatial localization tasks and got the highest scores in many benchmark datasets e.g. Kinetics, Charades, AVA, etc.

Our proposed ABN belongs to the category of two-stage TAD and boundary-based TAPG approach where our focus is human perception-based video feature extraction which aims to obtain semantic video representation followed by human perception of action understanding. 

\section{Our Proposed Method: Agent-Aware Boundary Network (ABN)}

\subsection{Problem Formulation}
Considering an untrimmed video $\mathcal{V} = \{x_l\}^{L}_{l=1}$ with $L$ frames, our goal is to generate a set of temporal segments which inclusively and tightly contain actions of interest. Given a set of ground truth action segments $\mathcal{A} = \{a_i=(s_i, e_i)\}^{M}_{i=1}$ having $M$ temporal segments with an action segment comprised of a starting timestamp ($s_i$) and an ending timestamp ($e_i$), our objective is formalized by minimizing the following objective function:
\begin{equation}
\mathcal{L}=\sum\limits_{i=1}^M \log p(a_i|\mathcal{V})
\label{eq:objective}
\end{equation}

As proposed by \cite{lin2018bsn} and \cite{bmn}, the above objective function may also be achieved indirectly by decomposing the action proposal generation problem into detecting starting and ending timestamps of every actions together with their duration, as formulated below:
\begin{equation}
\mathcal{L}=\sum\limits_{i=1}^M \log p(s_i|\mathcal{V}) + \log p(e_i|\mathcal{V}) + \log p(e_i - s_i|s_i)
\label{eq:objective2}
\end{equation}

%\subsubsection{Formulation for Video Representation}

Instead of representing the video frames separately as individual feature vectors, in our proposed framework, we divide the video $\mathcal{V}$ into $T=\left \lfloor{\frac{L}{\delta}}\right\rfloor$ non-overlapping snippets of $\delta$ consecutive frames. Each snippet of $\delta$ frames captures the motions taking place alongside the appearances, which are crucial in detecting the starting and ending timestamps for each action.

We denote $\phi$ as a feature extraction function which is applied to every $\delta$-frame snippet, the visual representation feature sequence $F$ of the entire video $\mathcal{V}$ can be defined as:
\begin{equation}
\begin{split}
F & =\{f_i\}^{T}_{i=1} =\{\phi(x_{\delta\cdot(i-1)+1},...,x_{\delta\cdot i})\}^{T}_{i=1}
\label{eq:funF}
\end{split}
\end{equation}

%By applying a function $\phi$ as a represent function to the video $\delta$-frame snippets, we have the below feature list:

%$$F=\{f_i=\phi(x_{\delta*(i-1)+1},...,x_{\delta*i})\}^{\left \lfloor{\frac{L}{\delta}}\right \rfloor}_{i=1}$$

In most of the previous works \cite{TCN, MSRA, Prop-SSAD, CTAP, SRG, lin2018bsn, liu2019multi, bmn, dbg}, function $\phi$ is simply defined as the extraction of a feature vector from a hidden layer of C3D Network \cite{C3D_3}, Two-Stream Network \cite{2_stream_1}, or SlowFast Network \cite{SlowFast} given a $\delta$-frame snippet. However, as stated in the above sections, this may cause insufficient or noisy information capture because actions and the agents, who create the actions, usually take place in small spatial regions of the video.

Hence, in this work, we propose a novel action proposal generation model in videos, named Agent-Aware Boundary Network, equipped with a new feature extraction mechanism for the function $\phi$ namely Contextual Agent-Aware Representation Network (described in Sec. \ref{subsec:representation}) to be able to incorporate information from both agents and the interaction between them and surrounding environment. Our proposed ABN with the Contextual Agent-Aware Representation Network can be developed on any backbone such as C3D Network \cite{C3D_3}, Two-Stream Network \cite{2_stream_1}, or the latest model of SlowFast Network \cite{SlowFast}. More details are discussed in section \ref{sub:CAE feature}.

%\subsection{Pipeline}
%\label{Overall}

%The pipeline of our Agent-Aware Boundary Network (ABN) is illustrated in Fig. \ref{proposed_network}. Given a video $\mathcal{V}$, we first divide it into  $T=\left \lfloor{\frac{L}{\delta}}\right\rfloor$ of non overlapping $\delta$-frame snippets. For the $i^{th}$ snippet, we then apply our Contextual Agent-Aware Representation Network (section \ref{subsec:representation}) to extract the contextual feature at both global and local levels $f_i$. By concatenating all the features from the snippets, we obtain the video visual representation $F =\{f_i\}^{T}_{i=1}$ which is fed into Boundary Generator Network (section \ref{subsec:BMN}) for proposing temporal segments containing actions. The whole pipeline is connected in a fully differentiable fashion to facilitate training and deploying phases.
% \begin{figure}[ht]
%   \begin{center}
%     \includegraphics[width=0.95\linewidth]{Figs/full_architecture_detailed.pdf}
%     \caption{An overview architecture for our proposed ABN for action proposal generation where Agent-Aware Representation Network is in Fig.\ref{video_representation} and described in section \ref{sub:CAE feature}.}
%     \label{proposed_network}
%   \end{center}
%   %\vspace*{-\baselineskip}
% \end{figure}
\begin{figure*}[ht!]
  \begin{center}
  \includegraphics[width=0.95\textwidth]{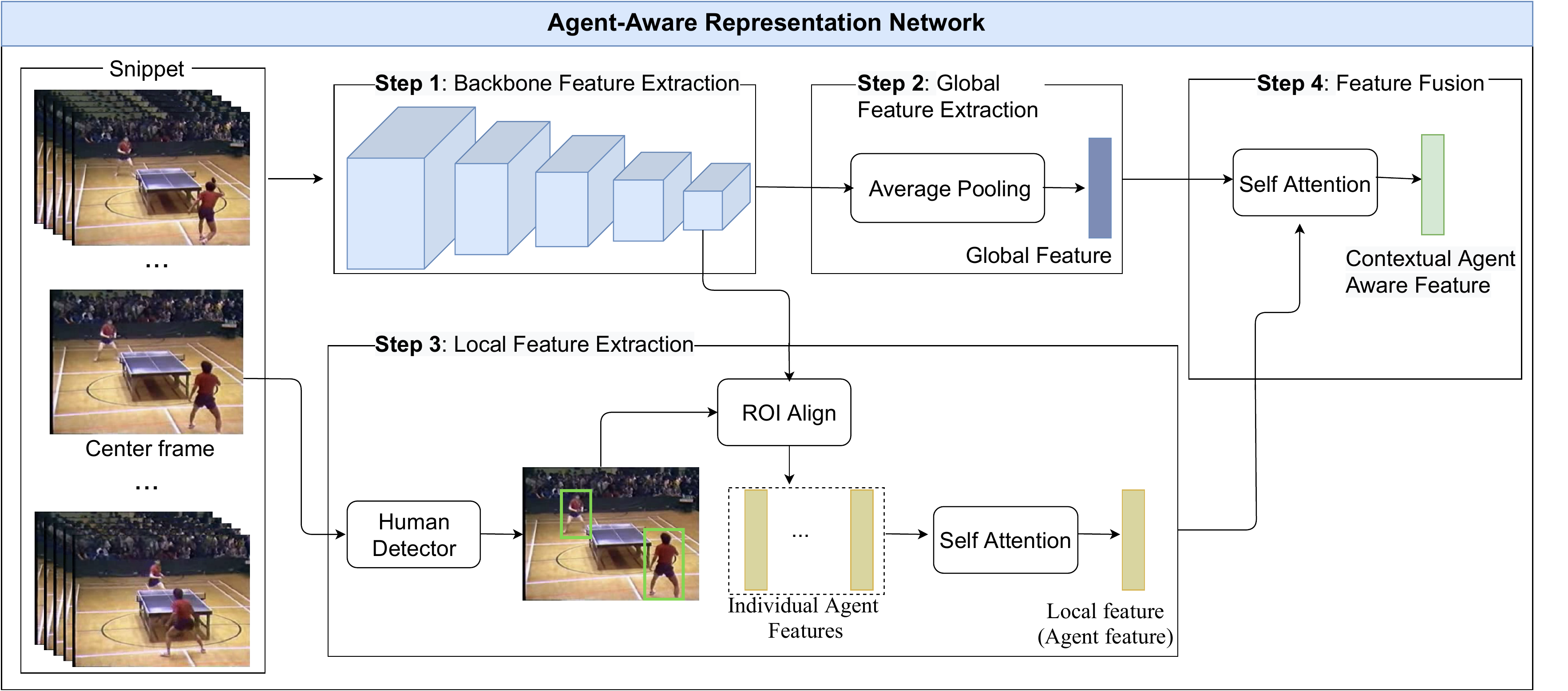}
  \caption{An overall architecture of our proposed contextual Agent-Aware representation network which contains four steps. Given a $\delta$-frame snippet, the final video visual feature is conducted by both global feature and local feature.}
  \label{video_representation}
  \end{center}
\end{figure*}

\subsection{Network Design}
\label{sub:CAE feature}
The ABN consists of two sub-networks and is demonstrated in Fig. \ref{fig:overall}. The first sub-network, \textbf{Contextual Agent-Aware Representation Network}, extracts contextual Agent-Aware visual representation of a $\delta$-frame snippet at both global and local levels and is detailed in section \ref{subsec:representation}.
The second sub-network, \textbf{Boundary Generation Network}, takes the first component as the input and generates the action proposals and is described in section \ref{subsec:BMN}.

\subsubsection{Contextual Agent-Aware Representation Network}
\label{subsec:representation} 
The contextual Agent-Aware representation network extracts contextual video visual representation of a $\delta$-frame snippet, which makes a significant contribution in TAPG. Considering our goal is extracting features for a $\delta$-frame snippet from frame $t$ to frame $t + \delta$, our Agent-Aware representation network is illustrated in Fig. \ref{video_representation} and consists of four steps, i.e. (i) backbone feature extraction, (ii) global feature extraction, (iii) local feature extraction, and (iv) feature fusion. Notably, there are two kinds of feature extracted in our proposed network, i.e., environment feature extracted through step 2 plays a role of global semantic level, and agent feature extracted through step 3 plays a role of local semantic level.

\textbf{Step 1: Backbone Feature Extraction:}
 The backbone network is used to encode global semantic information of the entire $\delta$-frame snippet. In order to prove the robustness of our proposed ABN, which is independent to the backbone network, we adopt Two-Stream, C3D \cite{C3D_3} and SlowFast \cite{SlowFast} networks in our experiments. These networks process a snippet of frames through multiple blocks of residual convolutional layers, with each block $B_i \in \{B_i\}_i^4$ produces a feature map $S_i \in \{S_i\}_i^4$. Assume $N$ is the last block of the backbone network, the feature map $S_N$ is then used as input for the two parallel pathways as in the next two steps.
 
 %C3D \cite{C3D_3} is a well-known model for resolving action recognition in videos. Thanks to its good performance and straightforward architecture, many works have utilized C3D as their backbone model for feature extraction, especially in temporal action proposals, hence, it is intuitive to reuse this network as a backbone model in our work for fair comparison with other works. Besides, SlowFast network \cite{SlowFast} has recently emerged to be the state-of-the-art model for action recognition in videos, therefore, we utilize SlowFast network in our experiment. 
 
 %this is a good opportunity to see if using SlowFast network \cite{SlowFast} as a backbone model could help our model to push up its results or not. %
% As shown in Fig. \ref{video_representation}, we take the output from the last block of backbone to attain a feature map $\Psi$. Then, the model splits into two parallel pathways, which are merged together at the end. These pathways, namely Environment and Agents pathways, are employed to extract the global information of overall scene i.e. environment and the local information of every human i.e. agent respectively.

\textbf{Step 2: Global Feature Extraction:}
To extract global feature, the feature map $S_4$ keeps going through average pooling and several fully connected layers until the second last layer, forming a vector which captures the overall scene, namely the global feature $\phi_e$. Because all spatial dimensions are processed, this pathway captures the abstract information at the global level of the scene, however, it may not able to capture the details like motions of agents inside.

\textbf{Step 3: Local Features Extraction:}
The local features extraction consists of two procedures. First, the local semantic feature vectors of each agent appearing in the video snippet are extracted. Then, all local feature vectors extracted from the first step are fused together to form an agent-aware feature vector. In order to fuse an arbitrary number of local semantic features, we employ a self-attention module with an average pooling layer, which is discussed below. 

For local semantic feature vectors extraction, we first detect agent appear in the $\delta$-frame snippet by a human detector. The center frame of the snippet is heuristically selected to feed into the detector because it is least diverged compared to frames at both ends of the snippet. We utilize Faster R-CNN \cite{FasterRCNN} model pre-trained on COCO dataset \cite{cocodataset} as our human detector after eliminating all object classes except the 'person' class. Detected human bounding boxes with confidence scores above 0.5 are then used to guide the RoIAlign \cite{MaskRCNN_ICCV17} to extract features from $S_4$, each feature storing local information about appearance and motion of the corresponding agent, called the local semantic feature.

After a set of local semantic features is formed, we employ a self-attention module to fuse them together into a single local agent-aware feature $\phi_a$. The self-attention module looks at the local semantic feature of each agent and assigns up-weights to agents who play important roles in the video snippet or are committing observed actions while assigning down-weights to the minor role agents.

In this step, the Faster R-CNN \cite{FasterRCNN} works as a hard attention module which eliminates all the background and only emphasizes humans or agents moving in the scene. On the other hand, the self-attention module works as a soft attention module which helps to concentrate on the right agents but also keeps information of the other agents because the activities we observe may require the interaction between these agents.

\textbf{Step 4: Feature Fusion:}
Finally, the environment feature $\phi_e$ and the agent aware feature $\phi_a$ are fused by another self-attention module (discussed below). While simultaneously processing these features, the self-attention module would re-weight them by a proper ratio, which helps the overall model to know which type of information to consider while reasoning the action proposals, i.e. deciding whether to emphasize on local information of the agents or global information of the scene.

\textbf{Self-Attention Module}
\label{subsec:trans_encoder}

In both TAPG and TAD, an arbitrary number of agents may appear in each snippet, which leads to difficulty to combine them into a single feature vector attentively to represent the snippet. Inspired by that problem, we propose a self-attention module which adopts the Transformer Encoder model \cite{attention_is_all_you_need} to learn to re-weight the importance of the semantic features set based on each of themselves and fuse them together by an average pooling operation.

%Transformer Encoder model \cite{attention_is_all_you_need}, which is equivalent to scaled dot-product attention, is a self-attention model which aims to softly select information from multiple input sources. In our proposed ABN, Transformer Encoder model plays an important role not only in concentrating on the proper agents, e.g. humans, detected in the center frame of a snippet, but also in determining whether to focus on the local information of agents or the global information of environment in that snippet.

The self-attention module is employed twice within a snippet, i.e. (i) encode the list of individual agent features to a single multi-agent feature and (ii) fuse both the environment feature $\phi_e$ and the multi-agent feature $\phi_a$ to a snippet feature $f$. Generally, a Transformer Encoder model will encode the set of input features $\Gamma=\{\eta_i\}_{i=1}^{\gamma}$ to three matrices of latent states, namely keys $K=(k_i=\theta_k(\eta_i))$, queries $Q=(q_i=\theta_q(\eta_i))$ and values $V=(v_i=\theta_v(\eta_i))$. Notably, in the agent feature extraction (step 2), $\gamma$ is equivalent to the number of agents, $\eta$ is corresponding to individual feature and $\Gamma$ is a single multi-agent feature whereas in the feature fusion (step 4) $\gamma$ is set as 2 and $\eta$ is corresponding to environment feature and multi-agent feature and $\Gamma$ is snippet feature. $\theta$ is defined as a fully connected layer. For each query state $q_i$ of an input feature, an attention function defined in Eq. \ref{attention} maps a query $q_i$  and a set of key-value pairs ($K$, $V$) to an output. The output is computed as a weighted sum of the values, where the weight assigned to each value is computed by a compatibility function of the query with the corresponding key as follows:

\begin{equation}
   \mathcal{A}(q_i, K, V) = \text{softmax}(\frac{q_i \cdot K^T}{\sqrt{d_K}})V
\label{attention}
\end{equation}

where $d_K$ is the number of dimensions in key states. Then, an average pooling layer is applied to fuse the resulting matrix $\Gamma_{\mathcal{A}}={\mathcal{A}(q_i, K, V)}_{i=1}^{\gamma}$ and form the overall context feature based on input features set.

The proposed self-attention model is utilized in our proposed ABN in a differentiable fashion and is trained along with the other parts of our network in an end-to-end way, hence, the resulting model may be able to properly generate the contextual Agent-Environment feature, which decreases the impact of background information in every snippet.

% As shown in Fig. \ref{video_representation}, a set of multiple humans $\mathcal{H}=\{b_h\}_{h=1}^H$ may be detected in the center frame leading to many corresponding features $\mathcal{F}_\mathcal{H}=\{f_h\}_{h=1}^H$ being extracted from the snippet. Hence, we apply a Transformer Encoder model to generate single contextual agents feature vector $f_\mathcal{H}$ to gate out semantically insignificant humans information as well as prioritize semantically important ones. Correspondingly, we utilize another Transformer Encoder model to fuse the environment feature with the mixed agents feature.
\begin{table*}[!t]
\centering
\caption{The detailed architecture of the boundary generation network which takes the contextual Agent-Aware visual feature $F$  as the input. $T$ and $D$ are the temporal length of the video and maximum duration of proposals in terms of number of snippets. The obtained outputs are $O_T$ and $O_P$, which are corresponding to boundary-predictions and proposal actionness scores.}

%\begin{tabular}{|p{0.3cm}|p{3.1cm}|p{2.2cm}|p{2.2cm}|}
\begin{tabular}{c|c|c|c}
\hline
\textbf{ID} & \textbf{Layer} & \textbf{Input}  & \textbf{Output} \\
\hline \hline
\multicolumn{4}{c}{\textbf{Base Module}} \\
\hline
1 & 1DConv.  \newline $256\times3/1$, ReLU & $I: F \times T$ & $O_1: 256\times T$\\
\hline
2 & 1DConv. \newline $128\times3/1$, ReLU & $O_1:256\times T$ & $O_2: 128\times T$\\
\hline
3 & 1DConv.  \newline $256\times3/1$, ReLU & $O_2:128\times T$ & $O_3: 256\times T$\\
\hline
\hline
\multicolumn{4}{c}{\textbf{Temporal Assessment Module (TAM)}} \\
\hline
4 & 1DConv.  \newline $2\times3/1$ , Sigmoid &  $O_3:256\times T$ & $O_T: 2\times T$\\
\hline
\hline
\multicolumn{4}{c}{\textbf{Proposal Assessment Module (PAM)}} \\
\hline
5 & Matching layer & $O_2:128\times T$ & $O_5:128\times32\times D \times T$\\
\hline
6 & 3DConv.  \newline $512\times32\times1\times1/(32,0,0)$ , ReLU &  $O_5:128\times32\times D\times T$ & $O_6:512\times 1 \times D \times T$\\
 \hline
7 & Squeeze   &  $O_6: 512\times 1 \times D \times T$ & $O_7: 512\times D \times T$\\
 \hline
8 & 2DConv.  \newline $128\times1\times1/(0,0)$ , ReLU &  $O_7:512\times D \times T$ & $O_8:128 \times D \times T$\\
 
 \hline
9 & 2DConv.  \newline $128\times3\times3/(1,1)$ , ReLU &  $O_8:128\times D \times T$ & $O_9:128\times D \times T$\\
 \hline
10 & 2DConv.  \newline $2\times1\times1/(0,0)$ , Sigmoid &  $O_9: 128\times D \times T$ & $O_P:2\times D \times T$\\
\hline
\end{tabular}
\label{tab:bmn}
\end{table*}

\subsubsection{Boundary Generation Network}
\label{subsec:BMN}

Our ABN belongs to the category of boundary-based approach and the boundary sub-network, i.e.  boundary generation network, contains three modules i.e. Base Module, Temporal Assessment Module (TAM) and Proposal Assessment Module (PAM). These modules are described follows:

\textbf{Base Module}
first processes the feature sequence $F$, which is extracted from the video by our contextual Agent-Aware representation network, through several 1D convolutional layers to extract temporal relationships between nearby snippet features. Those 1D convolutional layers are designed with a stride of 1 and same padding to reserve temporal length of the output feature sequence.

\textbf{Temporal Assessment Module (TAM)}
takes the features sequence from base module and estimates probabilities of every temporal location being a starting or ending boundary.

\textbf{Proposal Assessment Module (PAM)}
also takes the features sequence from base module and produces two matrices, each of which densely contains the confidence scores of every possible duration at every starting temporal point, but are trained by two different types of loss functions as suggested by \cite{bmn}. These matrices would have a shape of $D \times T$ with $D$ is the maximum length of the proposals in snippets that we consider and $T$ is the number of snippets. In this work, we set $D = T$ for ActivityNet-1.3 \cite{caba2015activitynet} and $D = T/2$ for THUMOS-14 \cite{THUMOS14} as suggested by \cite{bmn}.

Network architecture of Boundary Generation Network is given in Table \ref{tab:bmn}. In Table \ref{tab:bmn}, base module is represented by layer 1 to layer 3, temporal assessment module is represented by layer 4 and proposal assessment module is represented by layer 5 to layer 10.

%. The goal of Base Module is to handle the contextual Agent-Aware visual features for temporal relationship and provide a shared feature sequence for TAM and PAM. Base Module processes the input features through several 1D convolutional layers, which have stride of 1 to keep input features length, produces output features which are to be used by TAM and PAM simultaneously. On the one side, TAM has a job to produce the probabilities for every temporal point in the features set being a starting or ending boundaries. On the other side, PAM takes the features set from base module and produces two matrices, each of which densely contains the confidence scores of every possible duration at every starting temporal point, but are trained by two different types of loss functions as suggested by \cite{bmn}. these matrices would have a shape of $d \times t$ with $d$ is the maximum length of the proposals in snippets that we consider and $t$ is the number of snippets. in this work, we set $d = t$ as suggested by \cite{bmn}.

\subsection{Training Phase}

\subsubsection{Label Generation}
We follow \cite{bmn, lin2018bsn} to generate the ground truth labels for training process including starting labels, ending labels for TAM training and duration labels for PAM training.

The starting and ending labels are generated for every snippet of the video, which are called $L_S=\{l^s_n\}_{n=1}^T$ and $L_E=\{l^e_n\}_{n=1}^T$, respectively. The boundaries timestamps (starting and ending) of every action instance $a_i=(s_i, e_i)$ are rescaled into $T$-snippet range by multiplying them with $\frac{T \cdot \text{fps}}{L}$ where $\text{fps}$ is the frame rate of the video and the action instance $a_i \in \mathcal{A}$,  $\mathcal{A}=\{a_i\}_{i=1}^{M}$. After rescaling, the action instance $a_i$ becomes a new action instance $a^\delta_i=(s_i^\delta, e_i^\delta)$. For every snippet $t_n \in T$, we denote a temporal region $r_{n}=[t_n-\frac{1}{2},t_n+\frac{1}{2}]$. Analogously, for every pair of boundaries $(s_i^\delta, e_i^\delta)$ of action $a_i^\delta$, we denote regions $r^s_i=[s_i^\delta-\frac{3}{2}, s_i^\delta+\frac{3}{2}]$ and $r^e_i=[e_i^\delta-\frac{3}{2}, e_i^\delta+\frac{3}{2}]$ as their corresponding starting region and ending region. By this formulation, we have two sets of regions $R_S=\{r^s_i\}^M_{i=1}$ and $R_E=\{r^e_i\}^M_{i=1}$ for starting and ending boundaries, respectively. Finally, starting label $l^s_n$ and ending label $l^e_n$ of a snippet $t_n$ are calculated by the following functions:

\begin{multicols}{2}
\begin{equation}
\small
\nonumber
l^{\text{s}}_{n} = 
\begin{cases}
1,& \sum\limits^{M}_{i=1} \frac{r_{n} \cap r_i^{\text{s}}}{r_i^{\text{s}}} \geq 0.5
\\
0,              & \text{otherwise}
\end{cases}
\end{equation}

\begin{equation}
\small
\nonumber
    l^{\text{e}}_{n} = 
\begin{cases}
    1,&  \sum\limits^{M}_{i=1} \frac{r_{n} \cap r_i^{\text{e}}}{r_i^{\text{e}}} \geq 0.5 \\
    0,              & \text{otherwise}
\end{cases}
\end{equation}
\end{multicols}

The duration labels for a video are gathered into a matrix $L_D \in [0, 1]^{D \times T}$ where $D$ is the maximum length of proposals being considered in number of snippets, as suggested in \cite{bmn}, we set $D=T$ in all of our experiments. With an element at position $(t_i, t_j)$ stands for a proposal action $a_p=(t_s=\frac{t_j\cdot T}{t_v}, t_e=\frac{(t_j+t_i)\cdot T}{t_v})$, it will be assigned by $1$ if its Interaction-over-Union with any ground truth action in $\mathcal{A}=\{a_i\}_{i=1}^{M}$ reach a local maximum, or $0$ otherwise.

\subsubsection{Loss function}

As mentioned in section \ref{subsec:BMN}, TAM will generate probabilities vectors of starting and ending boundaries ($P_S \in \mathbb{R}^T$ and $P_E\in \mathbb{R}^T$), while PAM will generate two actionness scores matrices $P^{cc}_D \in \mathbb{R}^{D\times T}$ and $P^{cr}_D\in \mathbb{R}^{D\times T}$. These four outputs are trained simultaneously by different loss functions as following:
\begin{equation}
    \mathcal{L}_{TAM} = \mathcal{L}_{bin}(P_S, L_S) + \mathcal{L}_{bin}(P_E, L_E)
\end{equation}
\begin{equation}
    \mathcal{L}_{PAM} = \mathcal{L}_{bin}(P^{cc}_D, L_D) + \lambda_{reg} \cdot \mathcal{L}_{2}(P^{cr}_D, L_D)
\end{equation}
\begin{equation}
    \mathcal{L} = \lambda_1 \cdot \mathcal{L}_{TAM} + \lambda_2 \cdot \mathcal{L}_{PAM}
\end{equation}

As proposed by \cite{bmn, lin2018bsn}, we set $\lambda_{reg} = 10$ and $\lambda_1 = \lambda_2 = 1$, furthermore, $\mathcal{L}_{bin}$ is a weighted binary log-likelihood function to deal with imbalanced number of negative and positive examples in groundtruth labels. Generally, $\mathcal{L}_{bin}(\hat{Y}, Y)$ between prediction $\hat{Y} \in \mathbb{R}^N$ and groundtruth $Y \in \mathbb{R}^N$ is defined as follows:
\begin{equation}
     \frac{1}{N} \sum\limits^{N}_{i=1} \alpha^{+} \cdot Y_i \cdot \log{\hat{Y}_i} + \alpha^{-} \cdot (1 - Y_i) \cdot \log{(1-\hat{Y}_i)},
\end{equation}
where $\cdot$ is multiplication operator. The weighting parameters are automatically set by number of positives and negatives, specifically, $\alpha^{+} = \frac{N}{N^{+}}$ and $\alpha^{-} = \frac{N}{N^{-}}$, with $N$, $N^-$ and $N^+$ are total number of examples and total number of positive and negative examples, respectively.

\subsection{Inference Phase}
During inference, four outputs are generated by the boundary generation network from the features sequence extracted by our ABN, including $P_S$, $P_E$ from TAM output (output of layer 4 in Table \ref{tab:bmn}) and $P^{cc}_D$, $P^{cr}_D$ from PAM output (output of layer 10 in Table \ref{tab:bmn}). Peaking probabilities of starting and ending boundaries from $P_S$ and $P_E$, which are local maximums, are selected to form initial proposals by pairing every peak starting point with peak ending points behind them and within a pre-defined range. For a proposal formed by $t_s$ and $t_e$ boundaries with duration $d_p = t_e - t_s$, its score $s_p$, as proposed in \cite{bmn}, are computed as follows:
\begin{equation}
    s_{p} = P_S[t_s] \cdot P_E[t_e] \cdot \sqrt{P^{cc}_D[d_p, t_s] \cdot P^{cr}_D[d_p, t_s]}
\end{equation}
Then, with a list of proposals and their scores, a Soft-NMS \cite{SoftNMS} is applied to eliminate highly overlapped proposals before outputting the final list of proposals.

\begin{table*}[t]
\centering
\caption{Comparison in terms of AR@AN and AUC between our proposed ABN against other state-of-the-art \textbf{TAPG} methods on validation set and test set of \textbf{ActivityNet-1.3} dataset. The best performance is shown in \textbf{bold}. The second best performance is shown in \underline{\textit{italic}}}
\begin{tabular}{cccccc }
\toprule
Methods & Year & Feature   & \begin{tabular}[c]{@{}l@{}}AR@100 (val)\end{tabular} & \begin{tabular}[c]{@{}l@{}}AUC (val)\end{tabular} & \begin{tabular}[c]{@{}l@{}}AUC (test)\end{tabular} \\ \hline
TCN \cite{TCN}                &   ICCV2017       &     Two-Stream \cite{2_stream_1}      & -           & 59.58    & 61.56     \\ 
MSRA  \cite{MSRA}              &    CVPRW2017       &     P3D \cite{c3D_5}      & -           & 63.12    & 64.18     \\ 
SSTAD  \cite{SSTAD_BMVC17}      &    BMVC2017              &     C3D \cite{C3D_3}      & 73.01       & 64.40    & 64.80     \\ 
CTAP \cite{CTAP}                 &  ECCV2017      &     Two-Stream \cite{2_stream_1}      & 73.17       & 65.72    & -         \\ 
BSN   \cite{lin2018bsn}           &    ECCV2018         &     Two-Stream \cite{2_stream_1}      & 74.16       & 66.17    & 66.26     \\ 
SRG  \cite{SRG}                    &  TCSVT 2019  &     Two-Stream \cite{2_stream_1}      & 74.65       & 66.06    & -         \\ 
MGG \cite{liu2019multi}             &   CVPR2019        &      I3D \cite{i3d_2017}     & 74.54       & 66.43    & 66.47     \\ 
BMN   \cite{bmn}       &       ICCV2019         & Two-Stream \cite{2_stream_1}  & 75.01       & 67.10    & 67.19     \\ 
DBG   \cite{dbg}       &    AAAI2020             & Two-Stream \cite{2_stream_1} & \underline{\textit{76.65}}       & 68.23    & 68.57     \\ 
BSN++   \cite{BSN++}    &     ACCV2020   &     Two-Stream \cite{2_stream_1}     & 76.52       & 68.26    & -         \\
TSI++ \cite{tsi_accv}       &      ACCV2020        &      Two-Stream \cite{2_stream_1}      & 76.31    & 68.35 & 68.85         \\ 
MR~\cite{MR_eccv2020} & ECCV2020 & Two-Stream \cite{2_stream_1} & 75.27 & 66.51 & -- \\
SSTAP~\cite{wang2021self} & CVPR2021 & Two-Stream \cite{2_stream_1} & 75.20 &  67.23 & --\\
\hline
\multirow{3}{*}{\textbf{Our Proposed ABN}} & --  & Two-Stream      & 76.39   & 68.84 & \textbf{69.40} \\
 & -- & Slow-Fast & 76.64       & \underline{\textit{69.08}}    &     \underline{\textit{69.30}} \\ 
 & -- & C3D       & \textbf{76.72}     & \textbf{69.16}    & 69.26  \\ 
% &  & Two-Stream  & \textcolor{red}{NEED}          & \textcolor{red}{NEED}       & \textcolor{red}{NEED}        \\ 
% \multirow{3}{*}{ABN-DBG} & Slow-Fast & aaa         & aaa      & aaa       \\ \cline{2-5}
%                               & C3D       & aaa         & aaa      & aaa       \\ \cline{2-5} 
%                               & Two-Stream  & aaa         & aaa      & aaa       \\ \hline
\hline
\end{tabular}
\label{activitynet}
\end{table*}

\begin{table*}[t]
\centering
\caption{Comparisons with other state-of-the-art \textbf{TAPG} methods on testing set of \textbf{THUMOS-14} dataset in terms of AR@AN, where SNMS stands for Soft-NMS. The best performance is shown in \textbf{bold}. The second best performance is shown in \underline{\textit{italic}}}
\begin{tabular}{c| c c c c c c c }
\toprule
Feature  & Methods & Year & @50 & @100 & @200 & @500 & @1000              \\
\hline
\hline
\multirow{10}{*}{C3D}
& SCNN-prop \cite{anchor_2} &CVPR 2016 & 17.22 & 26.17 & 37.01 & 51.57 & 58.20        \\
& SST \cite{SST_CVPR2017} & CVPR 2017 & 19.90 & 28.36 & 37.90 & 51.58 & 60.27        \\
& TURN-TAP \cite{anchor_3} & ICCV 2017 & 19.63 & 27.96 & 38.34 & 53.52 & 60.75        \\
& BSN \cite{lin2018bsn} & ECCV 2018 & 29.58 & 37.38 & 45.55 & 54.67 & 59.48     \\
& MGG \cite{liu2019multi} & CVPR 2019 & 29.11 & 36.31 & 44.32 & 54.95 & 60.98        \\
& BMN \cite{bmn} & ICCV 2019 & 32.73 & 40.68 & 47.86 & 56.42 & 60.44     \\
& DBG+SNMS \cite{dbg} &  AAAI 2020 &   30.55 & 38.82 & 46.59 & 56.42 & 62.17     \\ 
& DBG \cite{dbg} &  AAAI 2020 &   32.55 & 41.07 & 48.83 & 57.58 & 59.55     \\ 
\cline{2-8}
%& \textbf{Our Proposed ABN} &  &\textbf{33.36}&\textbf{42.93}&\textbf{50.34}&\textbf{59.10}&\textbf{64.03}\\
%& \textbf{Our Proposed ABN} &  &\textbf{33.76}&\textbf{43.83}&\textbf{51.74}&\textbf{60.43}&\textbf{65.54}\\
& \textbf{Our Proposed ABN + SNMS} & -- & \underline{\textit{34.25}} & \underline{\textit{44.01}} & \underline{\textit{52.05}} & \textbf{60.57} & \textbf{65.39} \\
& \textbf{Our Proposed ABN + NMS} & --
& \textbf{36.01} & \textbf{45.41} & \textbf{52.74} & \underline{\textit{59.91}} & \underline{\textit{62.47}} \\
\hline
\hline
\multirow{9}{*}{Two-Stream}
& TURN-TAG \cite{anchor_3}     &  ICCV 2017 & 18.55 & 29.00 & 39.61 & - & -                \\
& CTAP  \cite{CTAP}  & ECCV 2018  & 32.49 & 42.61 & 51.97 & - & -                \\
& BSN \cite{lin2018bsn} & ECCV 2018 & 37.46 & 46.06 & 53.21 & 60.64 & 64.52          \\
& MGG  \cite{liu2019multi}  & CVPR 2019   & 39.93 & 47.75 & 54.65 & 61.36 & 64.06        \\
& BMN \cite{bmn}& ICCV 2019 &    39.36 & 47.72 & 54.70 & 62.07 & 65.49     \\
& DBG+SNMS \cite{dbg} &  AAAI 2020 &    37.32 & 46.67 & 54.50 & \underline{\textit{62.21}} & 66.40     \\ 
%& DBG \cite{dbg} & AAAI 2020 &    40.89 & 49.24 & 55.76 & 61.43 & 61.95     \\ 
& DBG+NMS\cite{dbg}   & AAAI 2020     & 40.89 & 49.24 & 55.76 & 61.43 & 61.95 \\
& SSTAP~\cite{wang2021self} & CVPR2021 & \underline{\textit{41.01}} &  \underline{\textit{50.12}} & \underline{\textit{56.69}} & -- &  \textbf{68.81} \\
% & TSI++\cite{tsi_accv} & ACCV 2020    & \underline{\textit{42.30}} & 50.51 & 57.24 & 63.43 & -     \\
\cline{2-8} 
& \textbf{Our Proposed ABN + SNMS} & -- &   40.87 & 49.09 & 56.24 &   \textbf{63.53}   &    \underline{\textit{67.29}}   \\ 
& \textbf{Our Proposed ABN + NMS} & -- &   \textbf{44.89}&   
\textbf{51.86}&  \textbf{57.36}   & 61.67 & 62.59\\
\bottomrule
\end{tabular}
\label{thumos}
\end{table*}

\begin{figure*}[ht]
  \begin{center}
    \includegraphics[width=0.75\linewidth]{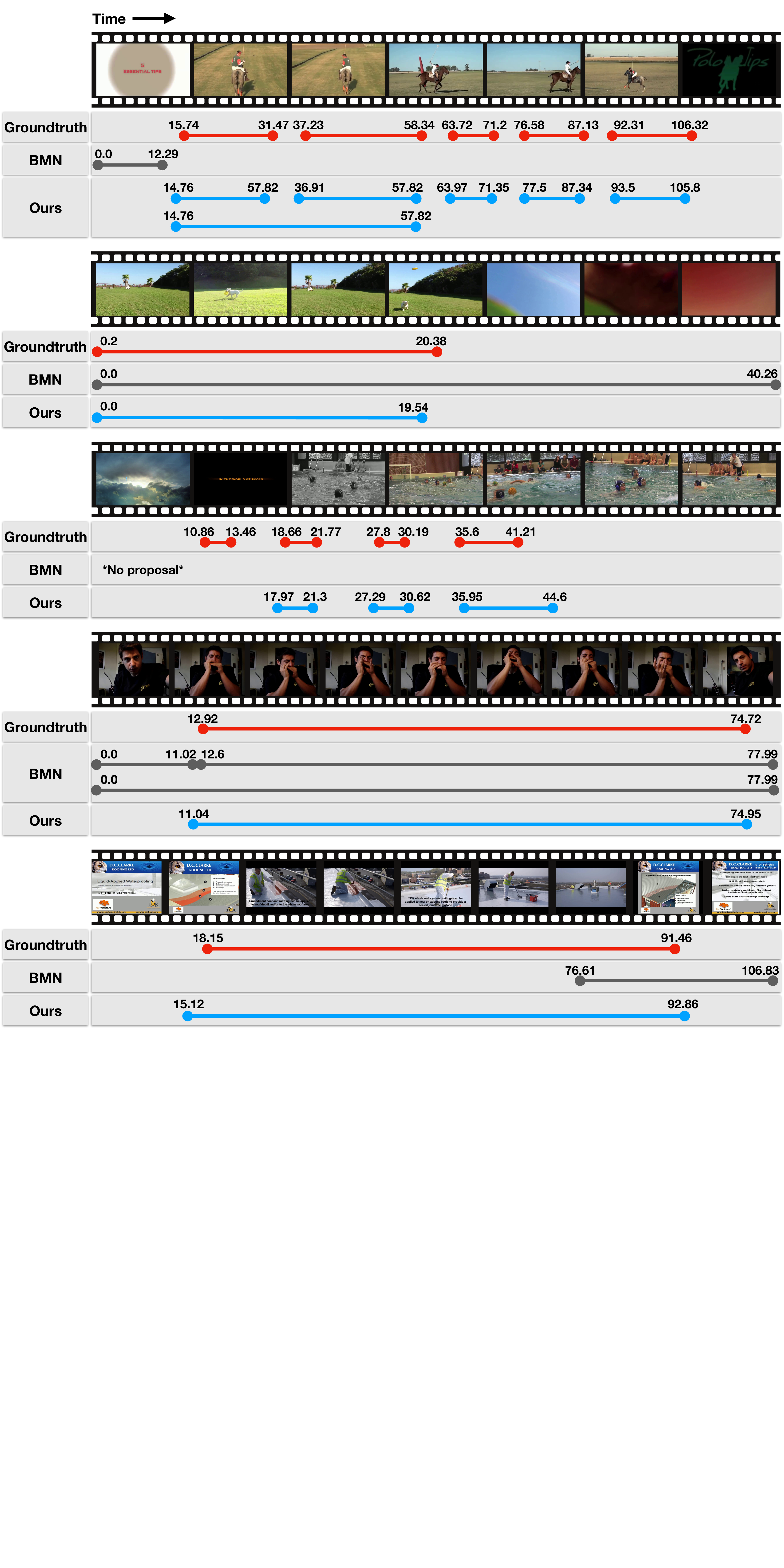}
    \caption{Qualitative results of proposals by BMN \cite{bmn} and our proposed ABN on ActivityNet-1.3 \cite{caba2015activitynet}, we use our best performed configuration which includes C3D \cite{C3D_3} as backbone feature extractor.}
  \label{qualitative_results_anet}
  \end{center}
  %\vspace{-10mm}
\end{figure*}

\begin{figure*}[ht]
  \begin{center}
    \includegraphics[width=0.75\linewidth]{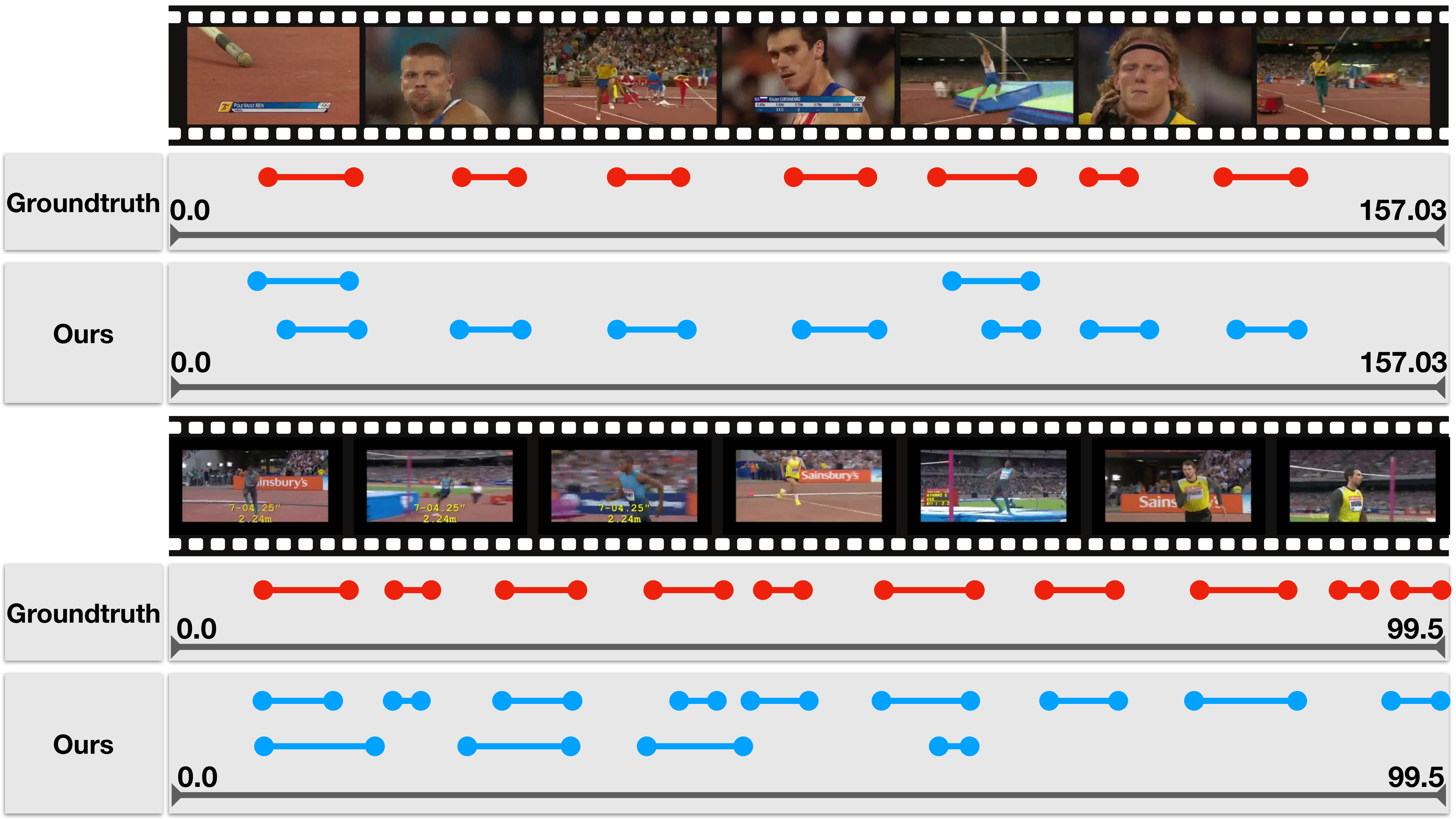}
    \caption{Qualitative results of proposals generated by our proposed ABN on THUMOS-14 \cite{THUMOS14}, we use our best performed configuration which includes Two-Stream \cite{2_stream_1} as backbone feature extractor.}
  \label{qualitative_results_thumos}
  \end{center}
\end{figure*}

\begin{table*}[t]
\centering
\caption{\textbf{Generalizability} evaluation on ActivityNet 1.3. The best performance is shown in \textbf{bold}.}
\begin{tabular}{c c c c c c}
&  & \multicolumn{2}{c}{Seen}   & \multicolumn{2}{l}{Unseen} \\ \cline{3-6}
Methods & Training Data & AR@100 & AUC   & AR@100  & AUC   \\ \hline
                         
\multirow{2}{*}{BSN \cite{lin2018bsn}}     & Seen+Unseen   & 72.40        & 63.80      & 71.84         & 63.99       \\ 
                         & Seen          & 72.42        & 64.02      & 71.32         & 63.38      \\ \hline
\multirow{2}{*}{BMN \cite{bmn}}     & Seen+Unseen   & 72.96        & 65.02      & 72.68         & 65.05       \\ 
                         & Seen          & 72.47        & 64.37      & 72.46         & 64.47       \\ \hline
\multirow{2}{*}{TSI++\cite{tsi_accv}}
& Seen+Unseen & \textbf{74.69} & 66.54 & 74.31 & 66.14 \\ 
& Seen        & 73.59 & 65.60 & 73.07 & 65.05 \\ \hline
\multirow{2}{*}{DBG\cite{dbg}}
& Un+/Seen & 73.30 &  66.57&  67.23 &  64.59\\ 
& Seen        & 72.95 &  66.23 &  64.77 &  62.18 \\ 
\hline

\multirow{2}{*}{\textbf{Our proposed ABN}} & Seen+Unseen   & 74.58       & \textbf{66.96 }     & \textbf{75.25 }        & \textbf{67.49}       \\ 
                         & Seen          & \textbf{74.40}        & \textbf{66.69  }    & \textbf{73.66}         & \textbf{65.49 }      \\ \hline
\end{tabular}
\label{tb:Generalizability}
\end{table*}

\begin{table*}[ht]
\centering
\caption{\textbf{TAD} results on ActivityNet-1.3 in terms of mAP@tIoU and average mAP, where our proposals are combined with video-level classification results generated by \cite{action_protocol}}
\begin{tabular}{l l l l l l l}
\hline
%& \multicolumn{4}{l}{Validation} & Testing \\ \hline
%Method & 0.5    & 0.75  & 0.95 & Average & Average \\ \hline
%CDC \cite{CDC}   & 43.83  & 25.88 & 0.21 & 22.77   & 22.90   \\ 
%SSN \cite{SSN}    & 39.12 & 23.48 & 5.49 & 23.98    &  28.28       \\ 
%BSN \cite{lin2018bsn}     & 46.45 & 29.96 & 8.02 & 30.03 &  32.87       \\ 
%BMN \cite{bmn}               & 50.07 & 34.78 & 8.29 & 33.85  & 36.42       \\ \hline
%\textbf{Our Proposed ABN  }  & 51.78 & 34.18 & 10.29 & 34.22  &   \textcolor{red}{NEED}     \\ \hline
%       & \multicolumn{4}{l}{Validation} & Testing \\ \hline
Method & & 0.5    & 0.75  & 0.95 & Average \\ \hline
CDC \cite{CDC} & CVPR2017  & 43.83  & 25.88 & 0.21 & 22.77   \\ 
%SSN \cite{SSN}    & 39.12 & 23.48 & 5.49 & 23.98   \\ 
BSN \cite{lin2018bsn} & ECCV2018 &  46.45 & 29.96 & 8.02 & 30.03 \\ 
BMN \cite{bmn}   & ICCV2019            & 50.07 & 34.78 & 8.29 & 33.85 \\
GTAD \cite{xu2020gtad} & CVPR2020   & 50.36 & 34.6 & 9.02 & 34.09 \\ 
P-GCN\cite{pgcn_cvpr2020}   & ICCV2019   & 42.9 & 28.1 & 2.5 & 27.0 \\
MR\cite{MR_eccv2020}  & ECCV2020      & 43.5 & 33.9 & 9.2 & 30.1 \\
BC-GNN~\cite{bai2020boundary} & ECCV2020 & 50.56 & 34.75 & 9.37 &  34.26 \\
TSI++~\cite{tsi_accv} & ACCV2020      & 51.2 & 35.0 & 6.6 & 34.2 \\
SSTAP~\cite{wang2021self} & CVPR2021& 50.72 & \textbf{35.28} & 7.87 &  \textbf{34.48} \\
\hline
\textbf{Our Proposed ABN  }  & -- & \textbf{51.78} & 34.18 & \textbf{10.29} & 34.22
\\
\hline
\end{tabular}
\label{action_detection_activitynet}
\end{table*}

\begin{table*}[ht]
\centering
\caption{Performance comparisons between our proposed ABN and the other proposal generation methods in terms of \textbf{TAD} on the testing set of THUMOS-14, where mAP is reported with tIoU set from 0.3 to 0.7 and Unet classifier is used}
\begin{tabular}{l l l l l l l} 
\hline
Method &  & 0.7   & 0.6  & 0.5   & 0.4   & 0.3 \\ \hline
% SST \cite{SST_CVPR2017}    & SCNN-cls  &  --  & 23.0 & -- & -- &  --   \\ 
% TURN-TAP\cite{SST_CVPR2017}  & SCNN-cls     &  7.7 & 14.6  & 25.6 & 33.2 & 44.1   \\ 
% BSN \cite{lin2018bsn}     & SCNN-cls & 15.0 & 22.4 & 29.4 & 36.6  & 43.1 \\ 
% BMN \cite{bmn} & SCNN-cls & 17.0 & 24.5 & 32.2 & 40.2 & 45.7 \\
% MGG \cite{liu2019multi} & SCNN-cls & 15.8 &  23.6 & 29.9 & 37.8 & 44.9\\ 
% \textbf{Our Proposed ABN  } & \textcolor{red}{NEED} & \textcolor{red}{NEED}& \textcolor{red}{NEED}& \textcolor{red}{NEED}& \tsdvextcolor{red}{NEED}& \textcolor{red}{NEED} \\
% \hline
\hline
SST \cite{SST_CVPR2017}    & CVPR2017  &  4.7 & 10.9 & 20.0 & 31.5 & 41.2 \\ 
TURN-TAP\cite{anchor_3}  & ICCV2017 & 6.3 & 14.1 & 24.5 & 35.3 & 46.3 \\
BSN \cite{lin2018bsn}     & ECCV2018 & 20.0 & 28.4 & 36.9 & 45.0 & 53.5 \\ 
BMN \cite{bmn} & ICCV2019 & 20.5 & 29.7 & 38.8 & 47.4 & 56.0 \\
MGG \cite{liu2019multi} & CVPR2019 & 21.3 & 29.5 & 37.4 & 46.8 & 53.9\\
DBG \cite{dbg} & AAAI2019 & 21.7 & 30.2 & 39.8 & 49.4 & 57.8\\
GTAN~\cite{gtan_cvpr2019}  & CVPR2019 & -- & -- & 38.8 & 47.2 & 57.8 \\
GTAD \cite{xu2020gtad} & CVPR2020 & 23.4 & 30.8 & 40.2 & 47.6 & 54.5\\
BC-GNN~\cite{bai2020boundary} & ECCV2020 & 23.1 & 31.2 &  40.4 & 49.1 & 57.1 \\

SSTAP \cite{wang2021self}& CVPR2021 & 22.8 & 32.8 & 42.3 & 51.5 & 58.4 \\
\hline
Our proposed ABN & -- & \textbf{25.56} & \textbf{37.04}& \textbf{46.12} & \textbf{53.95}&\textbf{59.87}  \\ % threshold 0.45, 3000 proposals
\hline
\end{tabular}
\label{action_detection_thumos}
\end{table*}

\begin{table*}[ht]
\centering
\caption{\textbf{Ablation studies} on the effectiveness of each component in the proposed ABN on ActivityNet-1.3 dataset with environment feature (Env.) and agent feature (Agent) in terms of AR@AN ($AN=100$) and AUC. The ablation study is conducted on various features i.e C3D, SlowFast and Two-Stream }
\begin{tabular}{l| l l l l l }
\hline
Feature &  & AR@1  & AR@10  & AR@100 & AUC \\ \hline
\multirow{3}{*}{\begin{tabular}[c]{@{}l@{}}C3D~\cite{C3D_3}\\\end{tabular}} 
& Env.      & 33.58 & 57.50 & 75.07 & 67.55 \\
% EnvOnly      & 33.01 & 58.69 & 76.40 & 68.80 \\
& Agent    & 30.14 & 53.80 & 72.76 & 64.54 \\
& Agent-Env. (Ours)             & \textbf{33.87} & \textbf{59.21} & \textbf{76.72} & \textbf{69.16} \\ \cline{2-6}
\hline\hline
\multirow{3}{*}{\begin{tabular}[c]{@{}l@{}}SlowFast~\cite{SlowFast}\end{tabular}}
& Env     & 33.74 & 57.11 & 75.59 & 67.85 \\
& Agent    & 32.24 & 52.77 & 72.67 & 64.27 \\
& Agent-Env. (Ours)               & \textbf{34.09} & \textbf{58.95} & \textbf{76.64} & \textbf{69.08}\\
\hline \hline
\multirow{3}{*}{\begin{tabular}[c]{@{}l@{}}Two-Stream~\cite{2_stream_1}\\\end{tabular}} 
& Env.      & 32.59 & 56.72 & 74.94 & 67.14 \\ 
& Agent    & 32.27 & 52.61 & 72.72 & 64.16 \\ 
& Agent-Env. (Ours)          & \textbf{33.61}  & \textbf{58.86} & \textbf{76.39} & \textbf{68.84}\\ 
\hline
\end{tabular}
\label{Ablation}
\end{table*}
\section{Experiments}
\subsection{Datasets \& Metrics}

\subsubsection{Datasets}
We evaluate our proposed method on two benchmark datasets, namely ActivityNet-1.3 \cite{caba2015activitynet} and THUMOS-14 \cite{THUMOS14}.

\noindent
\textbf{ActivityNet-1.3 }\cite{caba2015activitynet} is a large scale dataset for benchmarking methods in human activity understanding problems, in which, action proposals and action detection are the centers of attention. The dataset contains 200 distinct activity classes and a total of 849 hours of videos collected from YouTube. ActivityNet-1.3 \cite{caba2015activitynet} contains roughly 20K untrimmed videos which are divided into training, validation and test sets with the ratio of 0.5, 0.25 and 0.25, respectively. Each video in ActivityNet-1.3 \cite{caba2015activitynet} is annotated with one or more temporal intervals accommodating any activity out of 200 activities of interest. Due to the unavailability of annotations on test splits, we compare and report performances of our approach and other state-of-the-art methods on the validation set, unless otherwise stated.

\noindent
\textbf{THUMOS-14} \cite{THUMOS14}, on the other hand, is primarily a dataset for action recognition. Fortunately, a track of action localization and detection are derived from a portion of its videos. Concretely, 200 and 214 untrimmed videos are extracted from the validation and test sets of THUMOS-14 \cite{THUMOS14}, respectively, for training and testing methods in action detection.

\subsubsection{Metrics}
\label{subsubsec:metric}
To comprehensively evaluate the performance of the proposed ABN, we not only evaluate it in action proposals generation task, but also in action detection task.

For action proposals generation, on both ActivityNet-1.3 \cite{caba2015activitynet} and THUMOS-14 \cite{THUMOS14}, we measure AR with different Average Numbers (ANs) of proposals, denoted as AR@AN. AN is defined as the average number of proposals kept by every video in the dataset. Temporal intersection over union (tIoU) is used as the sole metric to classify a proposal. We follow the traditional practice, tIoU thresholds set from 0.5 to 0.95 with a step size of 0.05 are used on ActivityNet-1.3, while tIoU thresholds set from 0.5 to 1.0 with a step size of 0.05 are used on THUMOS-14. On ActivityNet-1.3 particularly, we report the score of area under the Average Recall (AR) versus Average Number of Proposals per Video curve (AUC), with the average number of proposals ranges from 0 to 100.

For action detection task, following previous works \cite{bmn, lin2018bsn}, we mainly evaluate our method by Mean Average Precision at tIoU (mAP@tIoU), with the tIoU in ranges $[0.5, 0.75, 0.95]$ and $[0.3, 0.4, 0.5, 0.6, 0.7]$ for ActivityNet-1.3 and THUMOS-14, respectively. At a specified tIoU, Average Precision is calculated for every action class and then averaged up to mAP@tIoU. On ActivityNet-1.3 particularly, we also report the Average mAP which is averaged among all mAP@tIoU scores.

For comparability purposes, we follow the same setting up which was described in \cite{bmn}. We re-scaled all videos to 1600 frames by linear interpolation and extracted features for every separate snippet with length $\delta = 16$ frames. Therefore, every video sequence will be represented by a feature sequence with the length of exactly 100 features.

%Due to the unavailability of annotations on test split, we compare and report performances of our approach and other state-of-the-art methods on the validation set, unless otherwise stated.

\subsection{Implementation Details}
On AcitivityNet \cite{caba2015activitynet}, we benchmark our proposed ABN on C3D \cite{C3D_3}, SlowFast \cite{SlowFast} and Two-Stream \cite{2_stream_2} backbones, the first two backbones are pre-trained on Kinetics-400 \cite{Kinetics} dataset, while the last one is pre-trained on recognition track of ActivityNet-1.3 \cite{caba2015activitynet}. The feature map $S_N$ of each backbone is extracted for the local feature extraction step, which has 2048 dimensions, 2304 dimensions and 400 dimensions for C3D, SlowFast and Two-Stream, respectively. We always keep this feature size through out our proposed network and output the Contextual Agent-Environment Feature of the same size as that of backbone feature.

On THUMOS-14 \cite{THUMOS14}, for fair comparisons with prior works \cite{lin2018bsn, bmn, dbg}, we employ C3D \cite{C3D_3} and Two-Stream \cite{2_stream_2} networks as the backbones of our proposed ABN. The C3D \cite{C3D_3} backbone is pre-trained on Kinetics-400 \cite{Kinetics}, whereas Two-Stream backbone is pre-trained on the action recognition track of ActivityNet-1.3 \cite{caba2015activitynet}. Output feature map $S_4$ from C3D and Two-Stream backbones are 2048 and 400 dimensions, respectively.

In the local feature extraction step, we adopt a Faster R-CNN \cite{FasterRCNN} model pre-trained on COCO dataset \cite{cocodataset} to detect human bounding boxes for generating local features later by RoI alignment with the feature map $S_4$.

The Transformer Encoders we used in Self-Attention Module for contextually merging local features into the local agent-aware feature or merging the local agent-aware feature with the global feature together share the same architecture of 4 attention heads and 1 transformer layer.

%We conduct experiments on ActivityNet-1.3 to compare our results with state-of-the-art methods. 
For every experiment, we trained our model for 10 epochs with initial learning rate of 0.0001 and Adam optimizer, the best performed model on validation set is chosen for further comparison.

In addition, we apply an augmentation where any groundtruth video, whose groundtruth actions having average length higher than a factor of $\tau_{upper}$ of its length, will be discarded. Contrarily, any groundtruth video, whose groundtruth actions having average length lower than a factor of $\tau_{lower}$ of its length, will be duplicated. We empirically observe that with $\tau_{upper}=0.98$ and $\tau_{lower}=0.3$, those augmentations during training will help the network achieve better performance and more robust on both datasets of ActivityNet-1.3 \cite{caba2015activitynet} and THUMOS-14 \cite{THUMOS14}.

%The detailed network architecture of the proposed ABN is given in Table \ref{tab:abn}. In this table, $F$ is the input feature dimensions. $T$ and $D$ are the temporal length of the video and maximum duration of proposals in terms of number of snippets. The obtained outputs are $O_T$ and $O_P$, which are corresponding to boundary-predictions and proposal actionness scores.

\subsection{Performance on TAPG}
\subsubsection{Compare with state-of-the-art methods}

Table \ref{activitynet} shows the comparison in terms of AR@AN (AN = 100) and AUC between the ABN against other state-of-the-art methods on both validation set and test set of ActivityNet-1.3 \cite{caba2015activitynet} dataset. Our performance is given in the last three rows of Table \ref{activitynet}. Compared against other state-of-the-art approaches, our proposed ABN obtains better performance on both AR@AN and AUC metrics regardless the backbone network. Concretely, the ABN outperforms BMN with 2.21\%, 2.11\% and 2.07\% in terms of AUC on test set, when using Two-Stream \cite{2_stream_2}, SlowFast \cite{SlowFast} and C3D \cite{C3D_3} backbones, respectively. With the most recent state-of-the-art, namely DBG \cite{dbg}, our proposed network makes the gaps of 0.83\%, 0.73\% and 0.69\% on AUC on testing set with Two-Stream, SlowFast and C3D backbones, respectively.

Additionally, Table \ref{thumos} summarizes the performances of our proposed ABN and other state-of-the-arts on testing set of THUMOS-14 \cite{THUMOS14} in terms of AR@AN (AN is in a set of [50, 100, 200, 500, 1000]). Our experiments are conducted on C3D \cite{C3D_3} and Two-Stream \cite{2_stream_2} backbone networks following previous works for fair comparisons. Besides, inspired by \cite{dbg}, we also measure the performance of our method with both NMS and Soft-NMS in post-processing phase. Suprisingly, our proposed ABN outperforms all the previous works with very large margins as shown in Table \ref{thumos}. We also noticed that using NMS will help the method to have better performances on top 200 proposals, while Soft-NMS will help the method to have better performances on more proposals e.g. 1000 ones.

Fig. \ref{qualitative_results_anet} illustrates some qualitative results of the generated proposals by ABN and BMN \cite{bmn} on ActivityNet-1.3 \cite{caba2015activitynet}. The experimental results show that Agent-Aware Boundary Network generates much better proposals, which almost perfectly cover the groundtruth events and tightly fit with their boundaries.

\subsubsection{Generalizability of Proposals}
One of the most important properties of a TAPG method is generating high quality proposals for unseen action categories. We follow the protocol defined in BSN \cite{lin2018bsn} and BMN \cite{bmn} to evaluate the generalizability of our proposed ABN. There are two un-overlapped action subsets: “Sports, Exercise, and Recreation” and “Socializing, Relaxing, and Leisure” of ActivityNet-1.3 are chosen as seen and unseen subsets separately. With such selection, there are 87 and 38 action categories, 4455 and 1903 training videos, 2198 and 896 validation videos on seen and unseen subsets separately. We first train our ABN on both seen training set and seen+unseen training set and then evaluate it on seen validation set and unseen validation set separately. Table \ref{tb:Generalizability} shows the performance of ABN along with the comparison with BSN \cite{lin2018bsn} and BMN \cite{bmn}. Compare with BSN and BMN, the ABN achieves better generalizability on both seen and unseen validation sets. This proves that our method can be used to generate proposals for activities and actions that it never met during the training phase.

\subsection{Performance on TAD}
Another important aspect worth considering is the utilization of proposals in action detection. Following BSN \cite{lin2018bsn} and BMN \cite{bmn} for a fair comparison, we adopt top-1 video-level classification results generated by method in \cite{action_protocol} on ActivityNet-1.3 to label the proposals generated by our method. Meanwhile, we use top-2 video-level classification results generated by UntrimmedNet \cite{untrimmetNet} to label proposals generated by our method on THUMOS-14. The labeled proposals are then evaluated on mAP@tIoU metric as described in Sec. \ref{subsubsec:metric}.

Table \ref{action_detection_activitynet} illustrates the performance of ABN and comparison with other state-of-the-art methods on ActivityNet-1.3 validation set. As we can see, our method outperforms BSN \cite{lin2018bsn}, BMN \cite{bmn} on all settings with a large margin and keeps a good distance with the most recent state-of-the-art mehtod in action detection named GTAD \cite{xu2020gtad} on all settings except mAP@0.75.

The experiment results on THUMOS-14 shown in Table \ref{action_detection_thumos} again emphasizes the superior performance of our ABN when compare with other methods including the state-of-the-art method in action detecion, namely GTAD \cite{xu2020gtad}.

\subsection{Ablation Study}

We conduct several ablation studies on the validation set of ActivityNet-1.3 dataset to analyze the contribution of individual feature in the ABN. In addition to different backbone networks, i.e. C3D \cite{C3D_3}, SlowFast \cite{SlowFast} and Two-Stream \cite{2_stream_2}, we have investigated the following ablation configurations for each backbone network.
%Concretely, we train two variations of our model with C3D and SlowFast backbone features separately:
\begin{itemize}
    \item Environment Feature Only (Env.): the network relies solely on global features to generate proposals.
    \item Agent Feature Only (Agent): the network relies completely on the contextual agent-aware features and does not use global feature.
    \item Both Environment and Agent Feature: both global feature and local feature are used and fused by Self-Attention Module. This network configuration is actually our proposed ABN.
\end{itemize}

Table \ref{Ablation} provides the results in terms of AR@AN (AN = 1, 10, 100) and AUC metrics of the ABN under different feature configurations on ActivityNet-1.3. In Environment Feature Only (Env.) configuration, there is only a global information about the environment of entire video frames, whereas in Agent Feature Only (Agent) configuration, there is only a local information about the agents. Table \ref{Ablation} has shown that each feature has its own contribution and the combination of both features exhibits the best performance. This result proves that our aforementioned observations are valid and reliable.

\section*{Conclusion}
In this paper, we proposed a novel contextual Agent-Aware Boundary Network (ABN) for the TAPG. Our ABN contains two components corresponding to Agent-Environment representation network and boundary generation network. The first component extracts the contextual visual representation of the video whereas the second component with boundary-based
mechanism aims at evaluating confidence scores of densely distributed proposals. 
%Our main contribution is mainly presented in the first component. 
Different from the previous works, which apply backbone network into the entire video frame, the video visual representation in the proposed ABN involves two parallel pathways: (i) the local pathway, which plays at the agents level and tells about where the agents are and what the agents are doing; (ii) the global pathway, which plays at a environment level and tells about how the environment affects after receiving the actions from the agents as well as the relationship between the agents, actions, and the environment. The experiments have demonstrated that our proposed ABN outperforms state-of-the-art proposal generation methods with C3D, SlowFast and Two-Stream backbone networks on both ActivityNet-1.3 and THUMOS-14 datasets. Our superior performance relies on both global feature and local feature, and demonstrates the robustness of the proposed ABN regardless of the backbone network as well as the effectiveness of our two-pathway contextual Agent-Environment visual representation. Additionally, our proposed method can also be generalized well to generate proposals for activities and actions that it never sees in training phase. Therefore, our method also shows the superior results in further applications like action detection task.

%\section*{Acknowledgment:}
%This material is based upon work supported by the National Science Foundation under Award No. OIA-1946391.
%
%The work is also supported by Vingroup Innovation Foundation (VINIF) in project code VINIF.2019. DA19, and  the Domestic Master/PhD Scholarship of VINIF.
%
%\section*{Disclaimer:}
%Any opinions, findings, and conclusions or recommendations expressed in this material are those of the author(s) and do not necessarily reflect the views of the National Science Foundation.

{\small
\bibliographystyle{ieeetr}
\bibliography{refs.bib}
}
\clearpage

\begin{IEEEbiography}[{\includegraphics[width=1in,height=1.25in,clip,keepaspectratio]{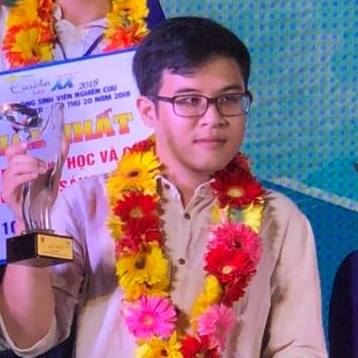}}]{} \textbf{Khoa Vo: } Khoa Vo is currently a PhD student in  Department of Computer Science and Computer Engineering at the University of Arkansas in Fayetteville. He received his B.S. degree in Computer Science from Honors Program, University of Science. From October 2019 to March 2020, he was an internship student in National Institute of Informatics, Japan. His research includes image captioning, object detection and temporal action detection. He has co-authored in publications appearing in conferences and journals including ICASSP, CVPR Workshops, Applied Sciences.
\end{IEEEbiography}

\begin{IEEEbiography}[{\includegraphics[width=1in,height=1.25in,clip,keepaspectratio]{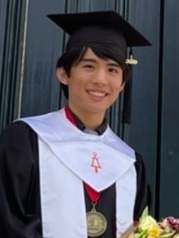}}]{} \textbf{Kashu Yamazaki: } He is currently a master student in Department of Computer Science and Computer Engineering at the University of Arkansas in Fayetteville. He received the B.S. with Summa Cum Laude in Mechanical Engineering from the University of Arkansas in Fayetteville in 2020. He is a member of Tau Beta Pi. His research interests includes objects detection and segmentation as well as video captioning. He has co-authored in publications appearing in conferences and journals including ICASSP, ICPR, and Artificial Intelligence Review.
\end{IEEEbiography}

\begin{IEEEbiography}[{\includegraphics[width=1in,height=1.25in,clip,keepaspectratio]{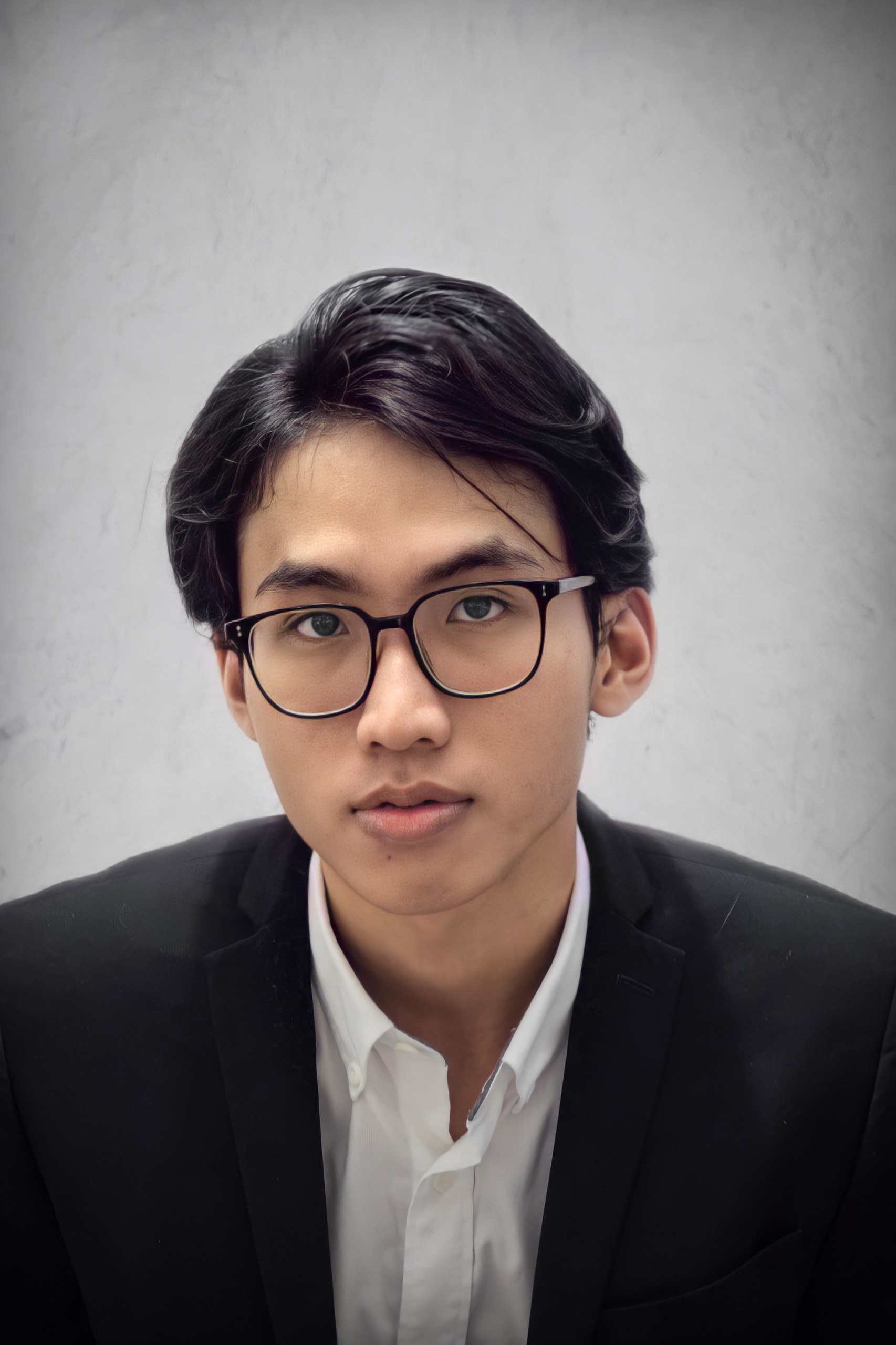}}]{} \textbf{Sang Truong: } Sang Truong is currently a PhD student in  Department of Computer Science and Computer Engineering at the University of Arkansas in Fayetteville. He received his B.S. degree in Automation and Control Engineering from International University, VNU-HCM in 2019. From September 2019 to May 2021, he worked in Finance and Machine Learning. His research interests includes Electrocardiography, Objects Detection and Actions Detection. 
\end{IEEEbiography}

\begin{IEEEbiography}[{\includegraphics[width=1in,height=1.25in,clip,keepaspectratio]{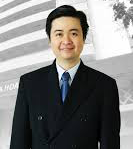}}]{} \textbf{Minh-Triet Tran:} Dr. Tran obtained his B.Sc., M.Sc., and Ph.D. degrees in computer science from University of Science, VNU-HCM, in 2001, 2005, and 2009. He joined the University of Science, VNU-HCM, in 2001. His research interests include cryptography and security, computer vision and human-computer interaction, and software engineering. He was a visiting scholar at National Institutes of Informatics (NII, Japan) in 2008, 2009, and 2010, and at University of Illinois at Urbana-Champaign (UIUC) in 2015-2016.
He is currently Head of Software Engineering Laboratory and Deputy Head of Artificial Intelligence Laboratory, University of Science, VNU-HCM. He is also the Deputy Head of Software Engineering Department, Faculty of Information Technology, University of Science, VNU-HCM. He was a member of the Executive Committee of the Information Security Program of Ho Chi Minh city. He is a member of the Management Board of Vietnam Information Security Association (South Branch) and also a member of the Executive Committee of ICT Program for Smart Cities (2018-2020) of Ho Chi Minh city.
\end{IEEEbiography}

\begin{IEEEbiography}[{\includegraphics[width=1in,height=1.25in,clip,keepaspectratio]{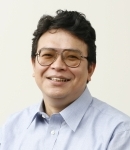}}]{} \textbf{Akihiro Sugimoto:} Dr. Sugimoto received his B.S, M.S, and Dr. Eng. degrees in mathematical engineering from the University of Tokyo in 1987, 1989, and 1996, respectively. He joined Hitachi Advanced Research Laboratory in 1989, and then temporally moved to Advanced Telecommunications Research Institute International (ATR), Japan in 1991. In 1995, he returned to Hitachi Advanced Research Laboratory where he lead a project on content-based image retrieval supported by the Ministry of International Trade and Industry in Japan. In 1999, he moved to Kyoto University as a lecturer at the Graduate School of Informatics. Since 2002, he has worked for NII, where he is currently a full Professor. From 2006 to 2007, he was also a visiting Professor at the University of Paris-Est Marne-la-Vallée, France. 

He has been an associate editor of International Journal of Computer Vision since 2014. He also served several top-tier conferences including IEEE International Conference on Computer Vision (ICCV), European Conference on Computer Vision (ECCV), Asian Conference on Computer Vision (ACCV), International Conference on Pattern Recognition (ICPR), and International Conference of 3D Vision (3DV) as an area chair, a program chair, and a general chair.  He is a regular reviewer of international conferences/journals in computer vision, AI, and pattern recognition. He has published more than 150 peer-reviewed journal/international conference papers.  He received Best Paper Awards from the Information Processing Society of Japan in 2001 and from the Institute of Electronics, Information and Communication Engineers (IEICE) in 2011. He is a member of IEEE and ACM.
\end{IEEEbiography}

\begin{IEEEbiography}[{\includegraphics[width=1in,height=1.25in,clip,keepaspectratio]{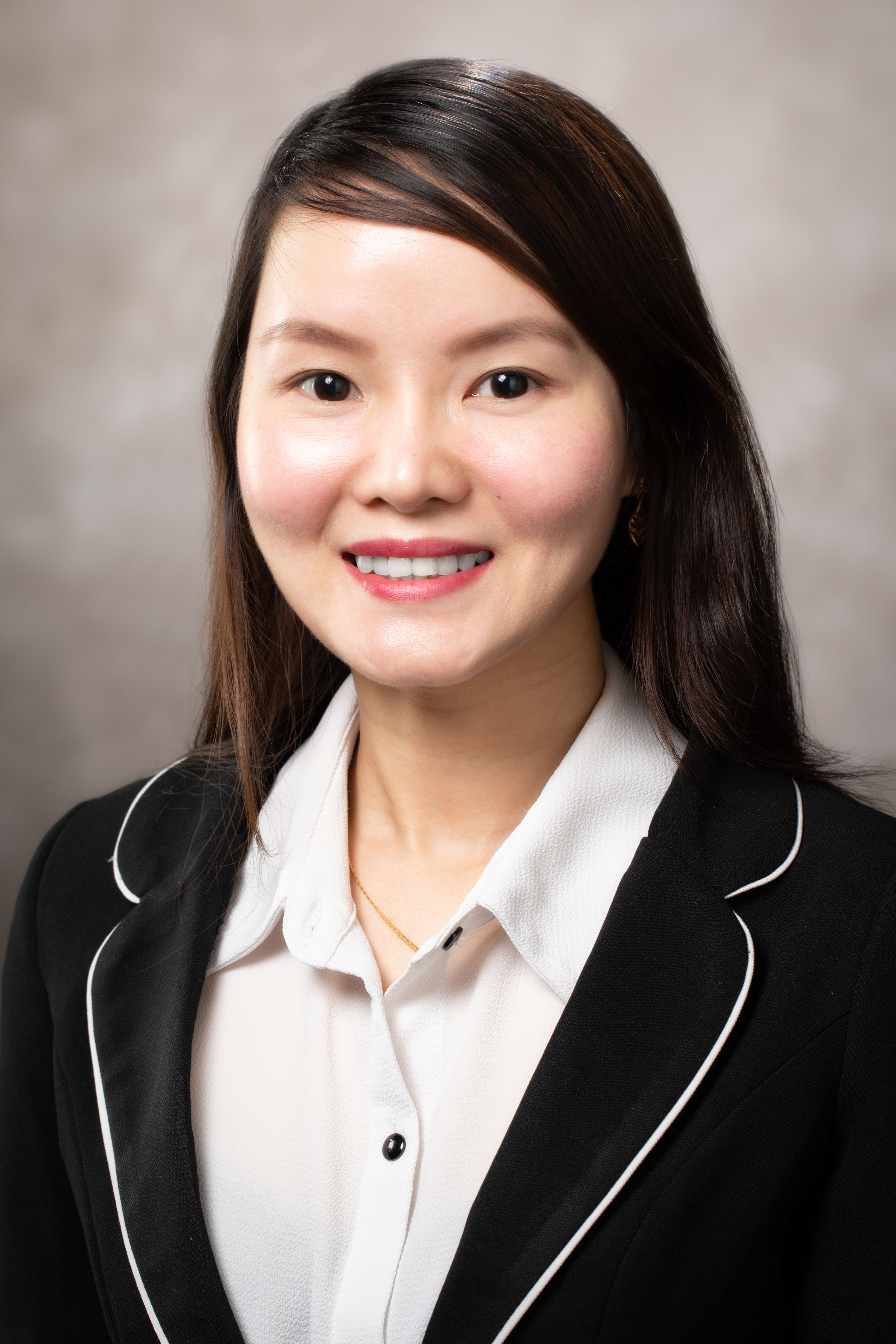}}]{} \textbf{Ngan Le:} Dr. Le is the director of Artificial Intelligence and Computer Vision lab and an Assistant Professor in the Department of Computer Science and Computer Engineering at University of Arkansas. She was a research associate in the Department of Electrical and Computer Engineering (ECE) at Carnegie Mellon University (CMU) in 2018-2019. She received the Ph.D degree in ECE at CMU in 2018, ECE Master degree at CMU in 2015, CS Master Degree at University of Science, Vietnam in 2009 and CS Bachelor degree at University of Science, Vietnam in 2005. Her current research interests focus on Image Understanding, Video Understanding, Computer Vision, Robotics, Machine Learning, Deep Learning, Reinforcement Learning, Biomedical Imaging, SingleCell-RNA.

Dr. Le is currently a guest editor of Scene Understanding in Autonomous (Frontier) and Artificial intelligence in Biomedicine and Healthcare (MDPI). She co-organized the Deep Reinforcement Learning Tutorial for Medical Imaging at MICCAI 2018, Medical Image Learning with Less Labels and Imperfect Data workshop at MICCAI 2019, 2020. Her publications appear in top conferences including CVPR, MICCAI, ICCV, SPIE, IJCV, ICIP etc, and premier journals including IJCV, JESA, TIP, PR, JDSP, TIFS, etc. She has co-authored 60+ journals, conference papers, and book chapters, 9+ patents and inventions. She has served as a reviewer for 10+ top-tier conferences and journals, including TPAMI, AAAI, CVPR, NIPS, ICCV, ECCV, MICCAI, TIP, PR, TAI, IVC, etc.
Website: \url{https://www.nganle.net}
\end{IEEEbiography}

\EOD

\end{document}